\documentclass{article} 

\usepackage{iclr2024_conference,times}
\usepackage{hyperref}
\usepackage{url}
\usepackage[utf8]{inputenc} % allow utf-8 input
\usepackage[T1]{fontenc}    % use 8-bit T1 fonts
\usepackage{hyperref}       % hyperlinks
\usepackage{url}            % simple URL typesetting
\usepackage{booktabs}       % professional-quality tables
\usepackage{amsfonts}       % blackboard math symbols
\usepackage{nicefrac}       % compact symbols for 1/2, etc.
\usepackage{microtype}      % microtypography
\usepackage{xcolor}         % colors
\usepackage{ulem}

\definecolor{omid_color}{RGB}{0 0 0}
\definecolor{removed_material}{RGB}{250 0 50}
\definecolor{unresolved}{RGB}{250 150 0}

\definecolor{navid_color}{RGB}{50 200 100}
\definecolor{revise_color}{RGB}{200 50 150}

\definecolor{lightgray}{gray}{0.9}

\usepackage{titletoc} % Needed for creating partial ToC

\usepackage[ruled]{algorithm2e} % For algorithms

\definecolor{comment_color}{RGB}{0 151 57}
\usepackage{mathtools}
\usepackage{colortbl}
\usepackage{caption}
\usepackage{wrapfig}
\usepackage{subfigure}

\iclrpreprintcopy 
\title{Large-Batch, Iteration-Efficient Neural Bayesian Design Optimization}

\author{%
  Navid Ansari\\
  Max Planck Institute for Informatics\\
  Saarbrücken, Germany \\
  \texttt{nansari@mpi-inf.mpg.de} \\
  \And
    Alireza Javanmardi\\
  Ludwig Maximilian University\\
  Munich, Germany \\
  \texttt{alireza.javanmardi@lmu.de} \\
  \And
  Eyke Hüllermeier\\
  Ludwig Maximilian University\\
   Munich, Germany \\
  \texttt{eyke@lmu.de} \\
  \And
    Hans-Peter Seidel\\
  Max Planck Institute for Informatics\\
  Saarbrücken, Germany \\
  \texttt{hpseidel@mpi-sb.mpg.de} \\
  \And
  Vahid Babaei\\
  Max Planck Institute for Informatics\\
  Saarbrücken, Germany \\
  \texttt{vbabaei@mpi-inf.mpg.de} \\
}

\begin{document}

\maketitle

\begin{abstract}

Bayesian optimization (BO) provides a powerful framework for optimizing black-box, expensive-to-evaluate functions. 
It is therefore an attractive tool for engineering design problems, typically involving multiple objectives.  
Thanks to the rapid advances in fabrication and measurement methods as well as parallel computing infrastructure, evaluating many design engineering problems can be heavily parallelized. 
This class of problems challenges BO with an unprecedented setup where it has to deal with very large batches, shifting its focus from sample efficiency to iteration efficiency. 
We present a novel Bayesian optimization framework specifically tailored to address these limitations. 
Our key contribution is a highly scalable, sample-based acquisition function that performs a non-dominated sorting of not only the objectives but also their associated uncertainties.  
We show that our acquisition function, in combination with different Bayesian neural network surrogates, is highly effective in extremely large-batch regimes with a minimal number of iterations. 
We demonstrate the superiority of our method by comparing it with state-of-the-art multi-objective optimizations.
We perform our evaluation on two real-world problems - airfoil design and 3D printing - showcasing the applicability and efficiency of our approach.
Our code is available at: https://github.com/an-on-ym-ous/lbn\_mobo

\end{abstract}

\section{Introduction} \label{intro}
% 1) application, importance and multi-objectivity 
% 2) offline vs online and missing in-between except single-objective 
% 3) ours 
Design of objects and materials that give us a specific \textit{performance}, typically defined by multiple objectives, is a long-standing, critical problem in engineering \cite{arnold2018directed}.
For real-world design problems, the forward mechanisms that govern the design processes are either sophisticated physics-based simulations or time- and labor-intensive lab experiments. 
We call these underlying mechanisms \textit{native} forward processes (NFP), which unlike \textit{surrogate} models, are the most faithful design evaluation tools at our disposal. 
% 
% \sout{An elegant solution to these design optimization problems is the \textit{online} sampling of the NFP through informed guesses about the best samples at each iteration.} % \cite{williams2006gaussian}. 
%
A powerful paradigm of design optimization is Bayesian optimization \cite{jones1998efficient} featuring a {surrogate} model that queries the NFP iteratively using a single data sample or a small batch of data. The choice of the next-iteration data is through optimizing a so-called \textit{acquisition function}. 

%%%%%%%%%%%%%%%{Problem}%%%%%%%%%%%%%%%%%%%%%%%  
%
{While Bayesian optimization literature focuses on solutions with a minimum number of NFP evaluations, a practically common but particularly underrepresented class of problems, especially in design optimization, is where intensive parallelization is feasible, but performing iterations is very demanding.}
A lower number of iterations is particularly beneficial in experimental scenarios where conducting lab experiments can be costly and time-consuming, making it desirable to minimize the number of lab visits.
Thanks to emerging high throughput experimentation \cite{macleod2022self}, many of these problems lend themselves to a large-batch setting where it is feasible to produce a large batch of samples in one iteration with an almost equivalent cost of evaluating a single sample. 
In such setups, it is desirable to have as large as possible batch sizes.
This type of setup is abundant in real-world applications, such as materials science \cite{raccuglia2016machine}, drug discovery \cite{dahl2014multi}, robotics \cite{marco2016automatic}, aerospace engineering \cite{chen2021mo}, manufacturing \cite{cucerca2020computational, panetta2022shape}, computational fluid dynamics \cite{jofre2022rapid, du2021rapid, sun2023physics}, etc. 
Despite the numerous real-world experiments that could gain from a large-batch optimization, there is a remarkable scarcity of Bayesian optimization algorithms adept at managing large batches, particularly for multi-objective optimization. 
% \sout{particularly those utilizing high-capacity, uncertainty-aware surrogate models like Bayesian Neural Networks (BNNs).}
%
To solve this class of problems effectively, we need to work toward shifting the paradigm from sample efficiency to iteration efficiency. 
Existing methods have limitations in retrieving good solutions either in a few iterations, or handling very large batches, or dealing with multiple objectives. 
%
% \sout{This might seem to be in contrast to the goal of traditional BO which is retrieving the best solutions with minimum function evaluation.
% %
% This ability to efficiently handle large data sets and reduce the number of iterations and lab visits showcases the significant utility of parallel large-batch BO } 
%

%%%%%%%%%%%%%%%{Solution}%%%%%%%%%%%%%%%%%%%%%%% 
% \vahid{paradigm shift from smaple efficiency to iteration efficiency}
We address these shortcomings by proposing a large-batch, neural multi-objective Bayesian optimization method (\textbf{LBN-MOBO}). 
%
% The larger batch size results in fewer iterations needed to identify the Pareto front, enhancing the efficiency significantly. 
%
Similar to any BO framework, our method has two key components. 
\textbf{First}, demonstrating the insufficiency of current acquisition functions in dealing with large batch regimes, we propose a highly practical acquisition function based on multi-objective sorting of samples 
where not only the performance objective but also its associated uncertainty is considered. 
By bringing in the uncertainty as an additional objective, LBN-MOBO can \textit{explore} previously unseen regions, preventing it from getting trapped in local minima.
The \textbf{second} component is a surrogate model capable of handling very large batches of data and also computing predictive uncertainty. We illustrate that, while our pipeline is compatible with all existing Bayesian neural networks (BNNs), Deep Ensembles (DE) \cite{lakshminarayanan2016simple} emerges as the method of choice, with the most balanced trade-off between performance and scalability.

We benchmark a range of state-of-the-art acquisition functions and surrogate models to show how current Bayesian optimizers struggle to exploit large batch sizes.
While we focus on evaluating \textit{neural} BO frameworks in the paper, suitable for learning via large batches, we provide extensive evaluation of a set of promising standard BO methods (relying on Gaussian processes) in the appendix.  
% %
% We then pick the most scalable surrogate model and propose a novel acquisition function and demonstrate how our pipeline efficiently exploits these large batch sizes and converge to a good solutions in significantly less iterations much more efficiently.
% 
Apart from testing our method with standard problems, we investigate two real-world problems one requiring hands-on lab work with a 3D printer and the second one an expensive CFD fluid dynamic simulation and show the Pareto front can be obtained with an order of magnitude less iterations. 
Our contributions include: 
\begin{itemize}    
    \item {A novel and scalable Bayesian optimization algorithm designed for the parallel, large-batch optimization of multi-objective problems, with a focus on iteration efficiency. Our method can retrieve a dense \textit{Pareto front} at each iteration which results in convergence in minimal iterations. Thanks to the large-batch capacity, it can be applied to problems with high-dimensional design spaces.} 
    
    \item A novel and practical acquisition function designed to effectively manage large batch multi-objective optimizations without inducing a computational bottleneck. The acquisition is embarrassingly parallelizable, shifting the computational bottleneck from the optimization algorithm to the computational infrastructure or experimentation capacities used to evaluate the NFP. Our acquisition is gradient-free, offering the flexibility to be paired with any arbitrary surrogate models. %, thereby broadening its applicability and utility in diverse optimization scenarios.
    
    \item A new benchmark of the state-of-the-art BO acquisition functions and surrogate models in large batch regimes.

    \item A variation of LBN-MOBO that is robust against irreducible noise (i.e., aleatoric uncertainty).
    % \item \navid{In Section~\ref{sec:LBN_MOBO_noise} we have presented a variation of LBN-MOBO that is robust against irreducible aleatoric noise.}
    
    % \item For a deeper efficiency evaluation, we have devised a novel algorithm for regret analysis of large batch, multi-objective evaluations (Section \ref{sec:regret}). Through a series of experiments, we demonstrate that the regret for LBN-MOBO consistently outperforms the counterpart optimizers.

    % \alireza{all the other contributions start with A...
    \item A novel algorithm for regret analysis of large batch, multi-objective evaluations (Section \ref{sec:regret}) for a deeper efficiency evaluation. Through a series of experiments, we demonstrate that the regret for LBN-MOBO consistently outperforms the counterpart optimizers.

    % (Figures \ref{fig:ZDT3_regret}, \ref{fig:regrets_DTLZ4}, \ref{fig:regrets_DTLZ1}, and \ref{fig:regrets_airfoil}).}
    % \item In this work, we adapt two real-world problems to the setting of large batch, low iteration optimization, and demonstrate the advantages of employing this approach. Utilizing our novel algorithm resulted in substantial savings in lab work effort in the first case. Furthermore, extensive parallelization significantly accelerated the computation in the latter scenario, showcasing the efficacy of our approach in diverse applications.
\end{itemize}

Given the breadth of the topics involved in this paper, we have prepared an extensive appendix that contains some key experiments and insights. We encourage the readers to refer to different parts of the appendix for further discussion.

\section{Related work: Multi-objective Neural Bayesian Optimization} \label{related}
% \vahid{@NN: Instead of following paragraph, say: 0) remind the large batch setup 
% 1) typical BO is GP but it doesn't scale (although we compare with the most promising ones)
% 2) That's why we focus on Neural BO that promise scalability }

Standard BO methods face two major bottleneck when given large batch sizes for multi-objective optimization.
Initially the acquisition function cannot scale with the data and becomes extremely slow and later the Gaussian process surrogate faces great difficulty in fitting the large amount of data.
Thus, in this section we focus on a variety of neural surrogate models for Bayesian optimization.
We start by reviewing suitable acquisition functions capable of handling multiple objectives for at least a batch size of two.  
In Section~\ref{sec:analysis}, we show how all of them can fail in a large batch setup. 
%

%
% In this section we introduce some of the main competitors and in Sections \ref{sec:analysis} and \ref{sec:evaluation} we evaluate their performance and analyze their limitations.
%
% For an effective comparison, the methods must at least possess two characteristics. First, they must be capable of managing multi-objective optimizations. Second, they must handle .
%
% \subsection{Multi-objective neural Bayesian optimization}
%
%%%%%%%%%%%%%%%%%%%%%%%%%%%%%%
\subsection{Multi objective batch acquisition functions} \label{sec:relatedWorkAcquisition}
\textbf{Expected Hypervolume Improvement (EHVI)} calculates the expected improvement in the hypervolume of the Pareto front. It has gained significant attention due to its capability to handle multi-objective optimization problems effectively. \citet{emmerich2005emo} initially proposed the concept of hypervolume improvement, and several advancements have been made since then.
\textbf{qEHVI} is a \textit{batch} version of EHVI, designed to make decisions about querying multiple points in the design space simultaneously \cite{daulton2020differentiable}. 
% \sout{It enhances the efficiency of EHVI by allowing parallel evaluations, which is particularly beneficial in scenarios where evaluations of the objective function are expensive or time-consuming.}
%

\textbf{Non-dominated EHVI (NEHVI)} is a variant of EHVI that focuses on the improvement of non-dominated points \cite{daulton2021parallel}. The introduction of NEHVI was a step forward in dealing with issues related to the scalability of EHVI by reducing the complexity from exponential to polynomial with respect to the batch size.
Moreover, NEHVI has demonstrated superior performance in addressing high-dimensional problems.
Following the development of qEHVI, \textbf{qNEHVI} emerged as the batch variant of NEHVI.

\textbf{Pareto Efficient Global Optimization (ParEGO)} transforms a multi-objective problem into a series of single-objective problems through scalarization functions, combining the strengths of Efficient Global Optimization (EGO) in a multi-objective setting \cite{knowles2006parego}.
The batch version of ParEGO, qParEGO, facilitates parallel evaluations of multiple points, significantly reducing the time required to find optimal solutions in multi-objective optimization scenarios \cite{daulton2020differentiable}. 

In Section \ref{sec:analysis}, we demonstrate the limitations of these acquisition functions when confronting larger batch sizes. 
%\sout{Specifically, many of these methods encounter significant challenges or inefficiencies as the batch size grows.} 
%
As we will see, several methods either fail to conclude the optimization process or encounter extreme inefficiencies for batch sizes exceeding 1000.
%

%%%%%%%%%%%%%%%%%%%%%%%%%%%%%%
\subsection{Bayesian neural surrogate models} \label{sec:relatedWorkSurroagate}
% \sout{When dealing with large datasets, Neural Networks (NNs) are well-suited and the most scalable candidate for modeling tasks.
% %
% As a result in the next subsection, we review the current state-of-the-art Neural network-based Bayesian optimizations.}
%
There have been numerous attempts to substitute Gaussian Processes (GPs) with neural networks to improve surrogate's scalability \cite{li2023study}. In order to convey uncertainty information, Bayesian neural networks are the key.
For inferring the posterior in a Bayesian neural network with stochastic parameters, several methods are available: 

\textbf{Hamiltonian Monte Carlo (HMC)} is a Markov Chain Monte Carlo (MCMC) method used for sampling from posterior distributions and has been recognized as a computational gold standard in Bayesian inference \cite{neal2011mcmc}. HMC leverages Hamiltonian dynamics to propose candidate states, reducing the correlation between consecutive samples and improving sampling efficiency.

\textbf{Stochastic Gradient HMC (SGHMC)} is a variant of HMC that incorporates stochastic gradients to scale to large datasets by working with mini-batches \cite{chen2014stochastic}. SGHMC addresses the challenges of noise introduced by mini-batch gradients, making it a scalable and robust approach for approximate Bayesian inference. 

\textbf{Deep Ensembles (DE)} trains multiple neural networks independently and aggregates their predictions to approximate the posterior predictive distribution \cite{lakshminarayanan2016simple}. 
This technique serves as a practical and effective heuristic for uncertainty estimation in BO.
   
\textbf{Monte Carlo Dropout (MC Dropout)} is a technique for approximating uncertainty in neural network models \cite{gal2016dropout}. It involves performing dropout at inference time and running multiple forward passes (Monte Carlo simulations) through the network, each time with different dropped-out nodes. By averaging the results of these passes, MC Dropout provides a measure of uncertainty associated with the model's predictions. 
Aside from methods for inferring the posterior of neural networks with stochastic parameters, several fundamentally different strategies exist for adapting a neural network as a Bayesian surrogate model:

\textbf{Infinite Width Bayesian Neural Networks (IBNNs)} are another class of neural surrogate models that can be seen as a bridge between the realm of neural networks and Gaussian processes.
    Research has shown that as a neural network's width approaches infinity, the distributions of the functions represented by the network converge to a Gaussian Process \cite{neal2012bayesian, lee2017deep}.
    This phenomenon implies that IBNNs can be seen as GPs with a specific neural network-derived covariance function.
    The  properties of IBNNs make them a scalable alternative to traditional GPs, especially in high-dimensional spaces.
    
\textbf{Deep Kernel Learning (DKL)} combines deep neural networks and Gaussian Processes to model complex and high-dimensional data sets, harnessing the representational power of deep learning and the uncertainty quantification of GPs \cite{wilson2016deep, wilson2016stochastic, ober2021promises}. In DKL, a neural network acts as a feature extractor, transforming input data into a feature space where a GP models the relationships between the transformed inputs and the output at the last layer. This approach offers advantages such as non-linear feature learning, uncertainty quantification, and flexibility in model architecture.

In the following sections, we will illustrate how various combinations of these surrogate-acquisition pairs struggle to process even a moderate batch size of 500 samples, rendering them unsuitable for our class of problems which typically involves batch sizes an order of magnitude larger.
%
% \sout{Furthermore, we will demonstrate that the primary bottleneck is the acquisition function.
% %
% In Section \ref{sec:acquisition}, a straightforward yet efficient acquisition strategy will be introduced that does not pose computational bottlenecks and is embarrassingly parallelizable.
% %
% Additionally, we will underscore that the Deep Ensembles approach emerges as the optimal surrogate model when balancing scalability with performance.} 
%
% In this regard, \cite{snoek2015scalable} applies adaptive basis function regression with a neural network to approximate uncertainty and uses neural Bayesian optimization for various single-objective optimizations. Similarly, \cite{durasov2021debosh} uses Deep Ensembles \cite{lakshminarayanan2016simple} and dropout \cite{gal2016dropout} to estimate uncertainty and applies the resulting Bayesian optimization on a graph neural network to generate designs with improved performance.
% %
% These methodologies all focus on a single objective and don't yield a Pareto front. One significant challenge, which we address in this work, is to devise an appropriate acquisition function that can generate a diverse set of Pareto front candidates without causing computational inefficiencies in the overall process. 

\section{Analysis of Contemporary Batch Neural Multi-Objective BO} \label{sec:analysis} 
In this section, we perform a critical evaluation of a set of advanced acquisition functions adept at multi-objective batch optimization.
Our findings highlight their limitations, specifically their inability to conduct optimization using large data batches.
In combination with the acquisition functions, we evaluate a range of contemporary neural surrogates.
The results indicate that the primary bottleneck in numerous scenarios is the acquisition function, followed by the surrogate.
For this empirical validation, we rely on the ZDT3 problem.
ZDT3 refers to one of the problems in the Zitzler-Deb-Thiele (ZDT) test suite \cite{zitzler2000comparison}, widely used to evaluate and compare the performance of multi-objective optimizations. ZDT3 specifically consists of two objectives and a disjoint Pareto front (Section~\ref{sec:ZDT_problems} of the Appendix). 
% The objective function as well as additional evaluations using other test suits are available in Section . 
% 

\begin{figure}
    \centering
    \includegraphics[width=0.8\textwidth]{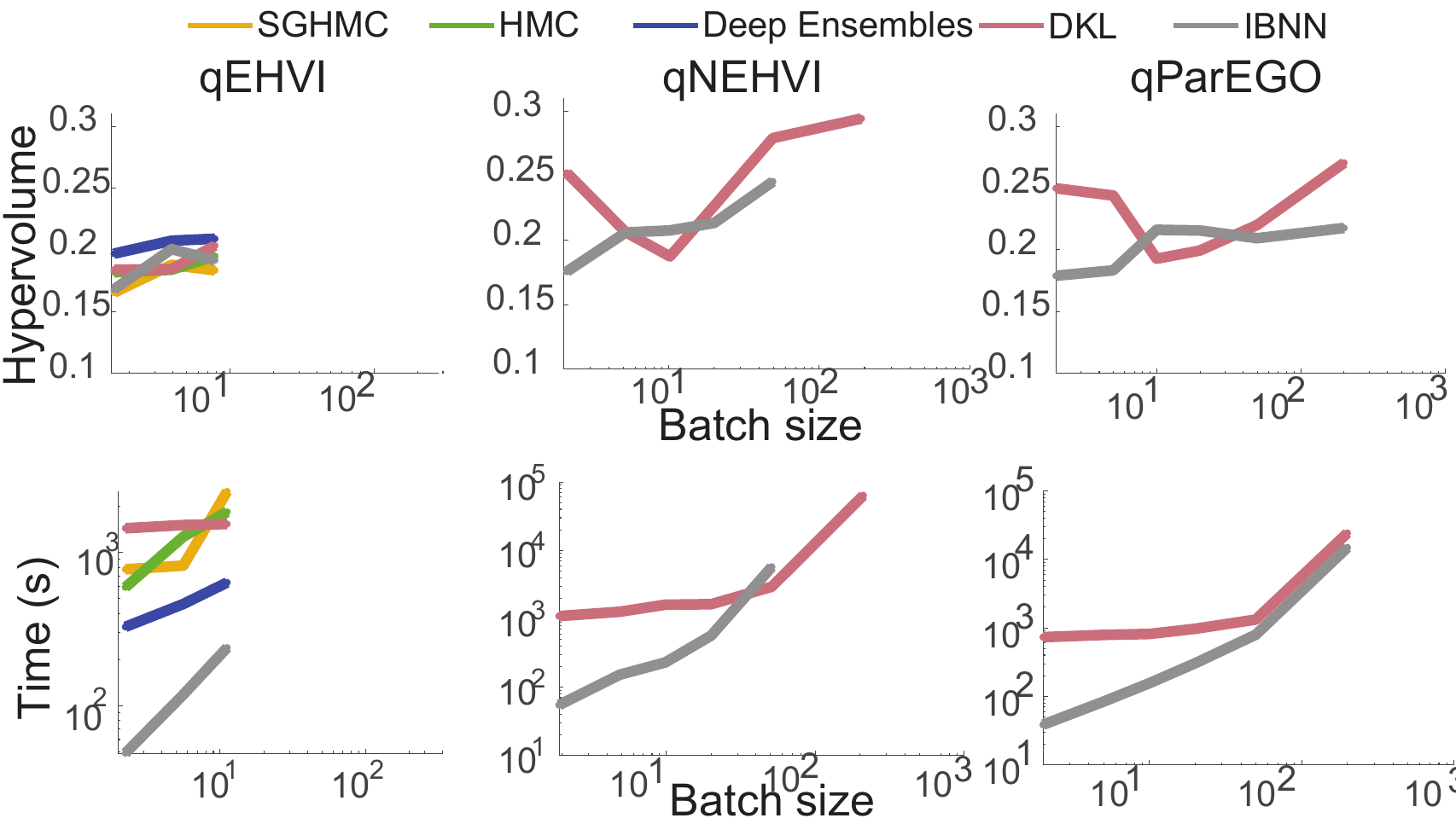}
    \caption{Optimizing 6D ZDT3 problem using a range of acquisition functions and surrogates.}
    \label{fig:NBOs}
\end{figure}

We start by computing the Pareto front of the 6-dimensional ZDT3 problem using qEHVI using several neural surrogates: DKL, HMC, IBNN, SGHMC, and Deep Ensembles (Section~\ref{sec:relatedWorkSurroagate}). For each batch size, the optimization is executed for 10 iterations, presenting the hypervolume of the best Pareto front achieved. Figure~\ref{fig:NBOs} (top left) demonstrates that when employing qEHVI as the acquisition function, the optimization stagnates at a batch size of 10. 
Notably, expanding to a batch size of 20 results in memory overflows independent of the surrogate.
The compute time for this experiment is shown in Figure~\ref{fig:NBOs} (bottom left).

For applying qNEHVI and ParEGO, we rely on BoTorch implementation that supports only DKL and IBNN. Figure~\ref{fig:NBOs} reveals that both acquisition functions can handle batch sizes larger than qEHVI.
Nonetheless, as depicted in Figure~\ref{fig:NBOs} (bottom row, middle and right), going beyond batch sizes larger than 200 increases the computational demand dramatically. 
 In this work, each algorithm's GPU run-time was restricted to 44 hours for every batch size optimization process. 
 More information regarding the hardware configuration is presented in Section \ref{sec:implementation_detail} of the Appendix.
Collectively, these experiments signify a gap in the capabilities of contemporary acquisition functions: They struggle  with batch sizes approaching 500 samples, a scenario frequently encountered in our real-world applications, as elaborated in Section~\ref{sec:evaluation}. Next, we introduce a novel acquisition function uniquely designed for working with extremely large batches.

% \begin{figure}
%          \centering
%     \includegraphics[width=0.4textwidth]{figs/NBOs.pdf}
%        \caption{Optimizing 6D ZDT3 problem using a range of acquisition functions and surrogates.}
%          \label{fig:NBOs}
% \end{figure}

\section{Method: Large-batch Neural MOBO} \label{method}

\begin{figure}
    \centering
        \includegraphics[width=1\textwidth]{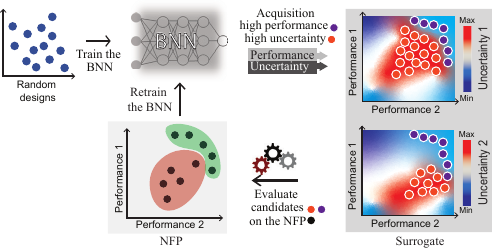}
    \caption{LBN-MOBO starts with training a Bayesian neural network ($f_{BNN}$) on random designs. We then run our acquisition function ($A\text{\tiny F}$) and compute a 2\textit{M}D Pareto front to explore promising (green) and under-represented regions (red) of the NFP. We then append the acquired candidates to the data set and retrain $f_{BNN}$. By incorporating uncertainty information alongside the Pareto front of the best performances (blue candidates),  we identify promising candidates in areas of high uncertainty, where there is potential for additional information (red candidates).}
    \label{fig:LBN-MOBO}
\end{figure}
% \begin{figure}
% \centering
% \subfigure[]{\label{fig:fig:LBN-MOBO}\includegraphics[width=0.50\textwidth]{figs/MONBO.pdf}}
% \subfigure[]{\label{fig:2MD_inset}\includegraphics[width=0.15\textwidth]{figs/2mD_pareto.pdf}}
% \caption{LBN-MOBO starts with training a Bayesian neural network ($f_{BNN}$) on random designs. We then run our acquisition function ($A\text{\tiny F}$) and compute a 2\textit{M}D Pareto front to explore promising (green) and under-represented regions (red) of the NFP. We then append the acquired candidates to the data set and retrain $f_{BNN}$. By incorporating uncertainty information alongside the Pareto front of the best performances (blue candidates),  we identify promising candidates in areas of high uncertainty, where there is potential for additional information (red candidates).}
% \label{fig:LBN_MOBO_and_2mD}
% \end{figure}
%%
Bayesian optimization for optimizing a black-box NFP, $\Phi$, uses a surrogate model fitted to an initial set of data from NFP to create a prior over the objective function. An acquisition function $A\text{\tiny F}$, derived from the surrogate model, guides the selection of the next samples, balancing exploration and exploitation. The surrogate model is updated by samples proposed by the acquisition function and evaluating by NFP. This process continues until a predetermined stopping criterion is reached.

Our method, LBN-MOBO, works on the same principles but is devised to achieve scalability. 
LBN-MOBO begins with a random sampling of the design space $\mathbf{U_{S}}(\mathcal{X})$ of the given NFP ($\Phi$). Subsequently, it fits an approximation of a Bayesian neural network surrogate $f_{BNN}$ to the randomly sampled dataset $\mathbf{X^{0}}$. The Bayesian neural network $f_{BNN}$ is capable of fitting to large data batches. 
Additionally, $f_{BNN}$, and particularly its approximation through Deep Ensembles (DE) \cite{lakshminarayanan2016simple}, enables computing predictive uncertainties ($\mathbb{F}_{\sigma}(\mathbf{x})$) in a fully parallelized manner (Section~\ref{sec:BNN}).
Upon training $f_{BNN}$, we utilize our acquisition function ($A\text{\tiny F}$) to compute the sample candidates, which explores both promising and under-represented regions (Section \ref{sec:acquisition}).
We append the calculated candidates to our data and utilize the updated dataset to train the BNN for the next generation.

Figure \ref{fig:LBN-MOBO} illustrates the stages of the LBN-MOBO algorithm using two objectives as an example.
Note that some of the candidates may not be positioned on the Pareto front of the NFP (indicated by the red regions), but they are still retained in the dataset. This is because they contribute to enhancing the information of $f_{BNN}$ and decreasing its uncertainty level ($\mathbb{F}_{\sigma}(\mathbf{x})$).
Algorithm \ref{alg:LBN-MOBO} provides a concise summary of all the steps of LBN-MOBO.

\begin{algorithm}[h]
\scriptsize % Adjust the font size
% \SetAlgoNlRelativeSize{-1}
% \SetAlgoLined
\textbf{Input}

$S$  \hspace{0.4cm}  \textcolor{comment_color}{\tcp{Batch size}}\
$Q$  \hspace{0.4cm} \textcolor{comment_color}{\tcp{Number of iterations of the main algorithm}}\
% $C$ \hspace{0.3cm} \textcolor{comment_color}{\tcp{Design space constraints}}\
$\mathcal{X}$ \hspace{0.25cm} \textcolor{comment_color}{\tcp{$\mathcal{X} \in \mathcal{R}^n$, $n$ dimensional design space}}\
$\Phi$  \hspace{0.3cm} \textcolor{comment_color}{\tcp{ Native Forward Process, e.g., a simulation}}\
% \textbf{Process parameters}
% $f_{BNN}(\mathcal{X})$ \hspace{0.35cm} \textcolor{comment_color}{\tcp{Bayesian neural surrogate}}\
\textbf{Output}
$P_{S}$ , $P_{F}$ \textcolor{comment_color}{\tcp{Pareto set(designs) and Pareto front(performances) of NFP}}\
\Begin{
    $\mathbf{X^{0}} \gets \mathbf{U_{S}}(\mathcal{X})$ 
      \textcolor{comment_color}{\tcp{Draw $S$ random samples from the design space.}}

    $\mathbf{Y^{0}}  \gets \Phi(\mathbf{X^{0}})$
        \textcolor{comment_color}{\tcp{Query $\Phi$ and form the data set.}}
    
    $data set \gets (\mathbf{X^{0}}$,\: $\mathbf{Y^{0}})$
    
    $f_{BNN}^{0} \xLeftarrow[\text{}]{\text{train}} data set$ \textcolor{comment_color}{\tcp{Train the BNN surrogate.}}     
     \For{$i\gets1$ \KwTo $Q$}{

      $P_{S}^{i} \gets A\text{\tiny F}(f_{BNN}^{i-1},\: S)$
    
      $ \; P_{F}^{i} \gets \Phi(P_{S}^{i})$  \textcolor{comment_color}{\tcp{Calculate the performance on the NFP.}}\
      $data set \gets (P_{F}^{i},\: P_{S}^{i})$
      \textcolor{comment_color}{\tcp{Append new data to the old.}}\
      $f_{BNN}^{i} \xLeftarrow[\text{}]{\text{train}} data set$  \textcolor{comment_color}{\tcp{Train the BNN surrogate.}}\
     }
\caption{Large-batch, neural multi-objective Bayesian optimization (LBN-MOBO).}
\label{alg:LBN-MOBO}
 }
\end{algorithm}
%
% \begin{algorithm}[h]
% \caption{Large-batch, Neural Multi-objective Bayesian Optimization (LBN-MOBO)}
% \label{alg:LBN-MOBO}
% \scriptsize % Adjust the font size
% \KwIn{
%     $S$ \tcp*{Batch size}
%     $Q$ \tcp*{Number of iterations}
%     $\mathcal{X}$ \tcp*{Design space $\mathcal{X} \in \mathbb{R}^n$, $n$-dimensional}
%     $\Phi$ \tcp*{Native Forward Process, e.g., a simulation}
% }
% \KwOut{
%     $P_{S}$, $P_{F}$ \tcp*{Pareto set (designs) and Pareto front (performances)}
% }
% \SetAlgoLined
% Initialize dataset: $\mathbf{X^{0}} \leftarrow \text{UniformSample}(\mathcal{X}, S)$\;
% Query $\Phi$: $\mathbf{Y^{0}} \leftarrow \Phi(\mathbf{X^{0}})$\;
% Initialize dataset: $\text{dataset} \leftarrow (\mathbf{X^{0}}, \mathbf{Y^{0}})$\;
% Train BNN surrogate: $f_{BNN}^{0} \leftarrow \text{Train}(dataset)$\;
% \For{$i = 1$ \KwTo $Q$}{
%     Select Pareto set: $P_{S}^{i} \leftarrow \text{AF}(f_{BNN}^{i-1}, S)$\;
%     Evaluate NFP: $P_{F}^{i} \leftarrow \Phi(P_{S}^{i})$\;
%     Update dataset: $\text{dataset} \leftarrow \text{dataset} \cup (P_{F}^{i}, P_{S}^{i})$\;
%     Train BNN surrogate: $f_{BNN}^{i} \leftarrow \text{Train}(dataset)$\;
% }
% \end{algorithm}

%
\subsection{Bayesian neural network surrogate} \label{sec:BNN}
Given that Deep Ensembles \cite{lakshminarayanan2016simple} presents the most balanced trade-off between performance and scalability (Section~\ref{sec:selection}), it will serve as the primary acquisition function in our pipeline. Here, we delve into its implementation and make slight modifications to enhance its performance further.
In this work, we employ a modified version of DE as an approximation of a BNN \cite{snoek2015scalable}.
DE consist of an ensemble of $K$ neural networks, $\hat{f}_{k}$, each capable of providing a prediction $\mu_{k}(\mathbf{x})$ and its associated \textit{aleatoric} uncertainty $\sigma_{k}(\mathbf{x})$ in the form of a Gaussian distribution $\mathcal{N}(\mu_{k}(\mathbf{x}), \sigma_{k}(\mathbf{x}))$.
DE has the unique advantage of separation between the epistemic and aleatoric uncertainty allowing us to use these information selectively in LBN-MOBO.

\paragraph{Aleatoric uncertainty $\mathbb{F}_{\sigma A}(\mathbf{x})$}
We assume irreducible aleatoric uncertainty $\mathbb{F}_{\sigma A}(\mathbf{x})$ is negligible in our problems. Hence, if we train DE carefully, we expect the predicted aleatoric uncertainty to be very small. 
However, in complex high dimensional problems it is very challenging to have a good estimation of  $\mathbb{F}_{\sigma A}(\mathbf{x})$ and it often predicts high aleatoric uncertainty where it is non-exist or fails to capture the existing ones.
This is a fundamental limitation of Bayesian neural networks when faced with complex problems. Thus, including aleatoric information can make LBN-MOBO unstable.
In Section \ref{sec:why_no_aleatoric} we provide further discussions and evaluation to elaborate on this issue.
As a result, under the assumption of negligible noise, we do not consider the aleatoric uncertainty $\mathbb{F}_{\sigma A}(\mathbf{x})$.
But, in Section \ref{sec:LBN_MOBO_noise}, we address the question of given an accurate estimation of $\mathbb{F}_{\sigma A}(\mathbf{x})$, can LBN-MOBO robustly handle the noise?
In Section \ref{sec:LBN_MOBO_noise}, we apply DE on simplified problems with severe noise where we ensure that DE is capable of approximating $\mathbb{F}_{\sigma A}(\mathbf{x})$ accurately.

\paragraph{Epistemic uncertainty $\mathbb{F}_{\sigma E}(\mathbf{x})$} 
To guide our optimizer to explore under-represented regions, we only require \textit{epistemic} uncertainty  $\mathbb{F}_{\sigma E}(\mathbf{x})$.
In the areas with higher epistemic uncertainty $\mathbb{F}_{\sigma E}(\mathbf{x})$ the networks in the ensemble fit differently due to a lack of information. Therefore, the regions with epistemic uncertainty are where we could potentially find better solutions. 
Thus, we only train $K$ neural networks using the traditional mean squared error (MSE) loss.
\begin{equation}
\mathcal{L}_{k}^{MSE} :=  (\mathbf{y^{*}} - \mu_{k}(\mathbf{x}))^{2}.
\label{eq:MSE_loss}
\end{equation}
Next, we extract the epistemic uncertainty $\mathbb{F}_{\sigma E}(\mathbf{x})$ from the networks in the ensemble:
\begin{subequations} \label{eq:mixture}
\begin{gather}
\mathbb{F}_{\mathbf{\mu}}(\mathbf{x}):= \frac{1}{K} \ \sum_{k} \ \mu_{k}(\mathbf{x}) \label{eq:F_mu}, \\
\mathbb{F}_{\sigma E}(\mathbf{x}) = \frac{1}{K} \  \sum_{k} (\mu_{k}^{2}(\mathbf{x}) - \mathbb{F}_{\mathbf{\mu}}^{2}(\mathbf{x})). \label{eq:F_sigma}
\end{gather}
\end{subequations}

{Apart from this modification, we find that providing a diverse set of activation functions across $K$ members of the ensemble significantly helps with obtaining higher quality uncertainty.
More details are provided in Section \ref{sec:implementation_detail} of the appendix.}

%%%%%%%%%%%%%%%%%%%%%%%%%%%%%%%%%%%%%%%%%%%%%%%
\subsection{2\textit{M}D acquisition function} \label{sec:acquisition}
% \sout{Here we show how epistemic uncertainty enables exploration in acquisition function ($A\text{\tiny F}$).} 
An acquisition function should predict the worthiest candidates for the next iteration of the Bayesian optimization \cite{shahriari2015taking}. This translates to not only selecting designs with high performance on the surrogate model but also considering the uncertainty of the surrogate model. Candidates in uncertain regions of the surrogate model may contain appropriate solutions and a powerful acquisition function should be able to explore these regions effectively. Several popular acquisition functions such as Expected Improvement \cite{jones1998efficient} and Upper Confidence Bound \cite{brochu2010tutorial} operate on this principle.

Without the loss of generality, we assume a problem that seeks to \textit{maximize} performance objectives. Our acquisition function employs the widely-used NSGA-II~\cite{deb2002fast} and specifically its multi-objective non-dominated sorting method (Section~\ref{sec:NDS}). This sorting is the key to find the Pareto front of the surrogate at a given iteration of LBN-MOBO. The main insight of our acquisition method is that instead of finding an $M$ dimensional Pareto front corresponding to $M$ objectives (each given by $\mathbb{F}_{\mathbf{\mu}}^{m}(\mathbf{x})$, $m \in \left[ 1, M \right]$), it finds a $2M$ dimensional Pareto front where $M$ dimensions correspond to performance objectives and the other $M$ dimensions correspond to the uncertainty of those objectives (each given by $\mathbb{F}_{\mathbf{\sigma E}}^{m}(\mathbf{x})$, $m \in \left[ 1, M \right]$). In other words, our acquisition function $A\text{\tiny F}$ \textit{simultaneously} maximizes the predicted objectives and their associated epistemic uncertainties (both given by our surrogate $f_{BNN}$): 
\small
\begin{subequations} \label{eq:acquisition}
\small
\begin{gather}
 \mathbb{F}(\mathbf{x}) = \mathbb{F}_{\mathbf{\mu}}^{m}(\mathbf{x}) \oplus \  \mathbb{F}_{\sigma E}^{m}(\mathbf{x}), \ m \in \left[ 1, M \right],\\
     % \text{ParetoFront} := \ \arg\max_{\mathbf{x}} \mathbb{F}(\mathbf{x}), \ \mathbb{F}(\mathbf{x}) \in \mathcal{R}^{2M}\\
     A\text{\tiny F} (\mathbb{F}(\mathbf{x}), S) := \text{ParetoFront}(\mathbb{F}(\mathbf{x}), S),
     \end{gather}
\end{subequations}
% \alireza{ 
% Let $\mathbb{F}(\mathbf{x}) 
% % = (\mathbb{F}_{\mathbf{\mu}}^1(\mathbf{x}), \ldots, \mathbb{F}_{\mathbf{\mu}}^m(\mathbf{x}),\mathbb{F}_{\sigma E}^1(\mathbf{x}), \ldots, \mathbb{F}_{\sigma E}^m(\mathbf{x})) 
% \in \mathcal{R}^{2M}$ represent the concatenation of all $2M$ objective functions. Our acquisition function returns the set of $S$ Pareto-efficient solutions that maximize the predicted objectives and their associated uncertainties simultaneously, i.e., 
% \begin{align*}
%     A\text{\tiny F} (\mathbb{F}(\mathbf{x}), S) := \text{ParetoFront}(\mathbb{F}(\mathbf{x}), S).
% \end{align*}
% }
where $\oplus$ represents the concatenation of $M$ prediction vectors and $M$ epistemic uncertainty estimation vectors and $\text{ParetoFront}(\mathbb{F}(\mathbf{x}), S)$ returns the set of $S$ Pareto-efficient solutions that maximize $\mathbb{F}(\mathbf{x})\in \mathcal{R}^{2M}$.
%
% And $\arg\max_{\mathbf{x}}$ operation is the process of finding the Pareto Front of this 2m dimensional problem using NSGA-II.
%
%
Note that in practice NSGA-II experiences a sample size bottleneck and struggles to scale effectively as the population expands. To overcome this limitation, we propose to compute in parallel independent acquisitions (different NSGA-II seeds) with smaller batch sizes, and combine the results. 
Ultimately, similar to our surrogate model, our acquisition function ($A\text{\tiny F}$) is fully parallelizable, and its performance remains unhampered even when batch size increases. Therefore, the sole limiting factor for executing LBN-MOBO is our parallel processing or experimentation capability when querying the NFP. 
\begin{figure}[h]  % Use 'h' to place it here
  \centering
  \includegraphics[width=20mm]{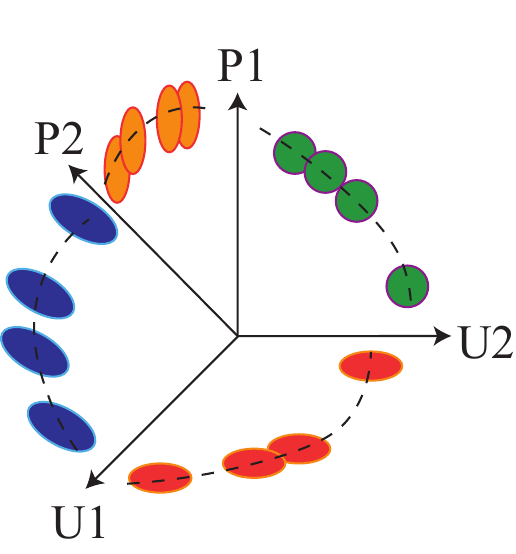}  % Adjusted width of the image
  \label{fig:2mD_pareto}
\end{figure}

The inset figure provides an intuitive explanation by showing a schematic four-dimensional acquisition function. Clearly, we are interested in evaluating the orange samples (currently measured only using the surrogate) on the NFP as they are suggested by $A\text{{\tiny F}}$ to be dominant in at least one \textit{performance} dimension (P1 or P2). 
On the other hand, the blue, red, and green samples are chosen partially or entirely due to their high uncertainty in at least one dimension (U1 or U2). These samples correspond to unexplored regions in the design space. They are beneficial in two ways: either they prove to be part of the Pareto front once being evaluated on the NFP, or they contribute to filling the gap between the surrogate and the NFP, leading to a more informative surrogate model.
This enhances the quality of the surrogate model, making it as similar as possible to the NFP, thereby improving its predictive power for subsequent iterations.
An ablation is provided in Section \ref{sec:uncertaimty-effect} to emphasize the importance of epistemic uncertainty for exploration.

\section{Evaluation and Discussion} \label{sec:evaluation}
Before starting with two challenging real-world problems we identify the most scalable surrogate capable of handling large batch sizes in a reasonable time.
While the focus of this section is on optimizing real-world problems using our neural Bayesian optimization, a comprehensive evaluation of various standard state-of-the-art Bayesian optimization techniques, as well as multi-objective evolutionary algorithms, can be found in Section \ref{sec:complementary_experiments} of Appendix.

\subsection{Selecting the most suitable surrogate} \label{sec:selection}
\begin{figure}
         \centering
        \includegraphics[width=1\textwidth]{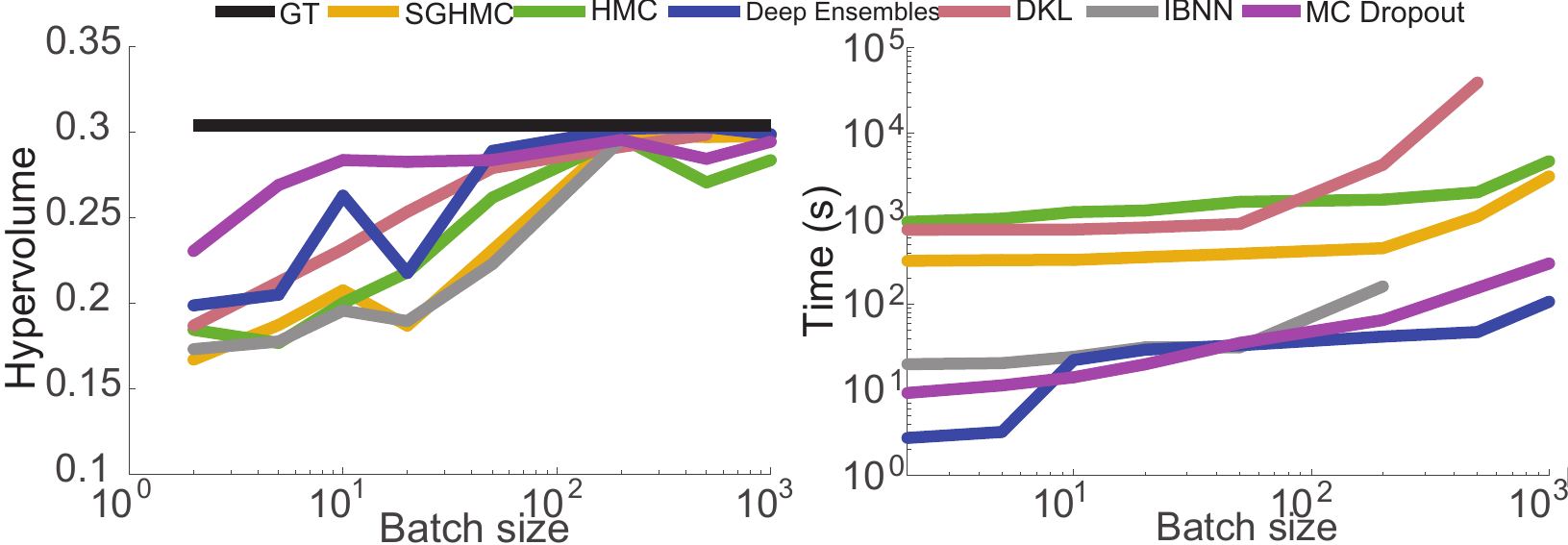}
       \caption{ZDT3 benchmark optimization via 2\textit{M}D acquisition function and various neural BO surrogate models. The fitting time of each model was recorded to assess computational efficiency.}
         \label{fig:2MD_time_hv}
\end{figure}

A pivotal aspect, distinct from the acquisition function selection, is the choice of the neural surrogate. 
Even though our pipeline can work with any surrogate models (Appendix Section \ref{sec:BNNs_2MD}), our objective is identifying the model that has the most efficiency and scalability. 
Analogous to Section~\ref{sec:analysis}, we assess the performance of various surrogate models, this time employing our 2\textit{M}D acquisition function.
As depicted in Figure \ref{fig:2MD_time_hv} (left), by using our proposed acquisition function, the bottleneck associated with the acquisition function is entirely alleviated, and all methods, with the exception of IBNN, have successfully completed the optimization for batch sizes up to 1000.
Figure~\ref{fig:2MD_time_hv} (right) shows the optimization time for different surrogates, each undergoing a 10-iteration optimization across a spectrum of batch sizes.
DE and MC dropout prove to be the most time-efficient models, adeptly conducting optimizations for batch sizes up to 1000 and finding the best Pareto front in the least number of iterations. These experiments substantiate DE and MC dropout as the clear choice to be paired with our novel surrogate model. 
{A detailed explanation of how we use the MC dropout in combination with our 2\textit{M}D acquisition is provided in the Appendix Section \ref{sec:MCdropout}.}
We have provided more challenging experiments in the appendix with larger input and output dimensions  (Appendix Section \ref{sec:DTLZ5}).
%

% With this successful combination, we are now ready to address real-world problems characterized by batch sizes an order of magnitude larger in the subsequent section.
%

%%%%% AB%%%%%%
\subsection{Real-world experiment set up} \label{sec:Experiments_setup_short} 
As discussed earlier, we found LBN-MOBO to be useful in two important classes of experiment: those involving cumbersome lab work or expensive simulations which can be parallelized.

\textbf{3D printer's color gamut experiment} is an examples of type 1 experiments.
The color gamut is the range of all colors that a device, such as a printer, can produce. 
The colorfulness is quantified using the CIE a*b* color space \cite{CIE2004}.
In this space the diversity of colors directly translates into the \textit{area} inside the contour of the CIE a*b* plot. The Bayesian optimization iteratively enlarges this area as it discovers more saturated colors \footnote{For this problem, we solve four Bayesian optimizations for four quadrants in order to advance the Pareto front in four different segments.}. 
3D printer is capable of generating many small (e.g., 1x1 mm) patches of color in a single operation making it a perfect show case for the benefits of large-batch optimization that LBN-MOBO can perform. 
In Sections \ref{sec:real_world} and \ref{sec:complementary_experiments} we show how this approach outperforms all other algorithms.
More details of this experiment can be found in Section \ref{sec:experiment_AB} of Appendix.

\textbf{Airfoil's} lift ($C_{l}$) and drag ($C_{d}$) coefficient optimization falls into the second category of problems.
In this setup we have control over the shape of the airfoils through 6 design parameters.
The goal is to optimize the the designs to generate the largest $C_{l}$ and $C_{l}/C_{d}$ ratio.
The evaluation of a single shape using a CFD simulator, i.e., NFP, is lengthy. Thus, to minimize the total optimization time we can rely on parallel computing to create large batches of data and find the best Pareto front in a few iterations.
More details of the airfoil experiment can be found in Section \ref{sec:experiment_airfoil} of Appendix.
\subsection{LBN-MOBO for real-world problems} \label{sec:real_world}
%

% \begin{figure}
%      \centering
%      \begin{subfigure}
%          \centering
%          \includegraphics[width=0.22\textwidth]{figs/Progress_hv_airfoil_dropout_ensemble.pdf}
%          \caption{}
%          \label{fig:Progress_hv_airfoil_dropout_ensemble}
%      \end{subfigure}
%      \hfill
%      \begin{subfigure}
%          \centering
%          \includegraphics[width=0.22\textwidth]{figs/comparison_airfoil_dropout_ensemble.pdf}
%          \caption{}
%          \label{fig:comparison_airfoil_dropout_ensemble}
%      \end{subfigure}
%      \hfill
%      \begin{subfigure}
%          \centering
%          \includegraphics[width=0.22\textwidth]{figs/Progress_hv_AB_dropout_ensemble.pdf}
%          \caption{}
%          \label{fig:Progress_hv_AB_dropout_ensemble}
%      \end{subfigure}
%      \hfill
%      \begin{subfigure}
%          \centering
%          \includegraphics[width=0.22\textwidth]{figs/Gamut_AB_44ink_dropoit_ensemble.pdf}
%          \caption{}
%          \label{fig:44_ink_gamut_dropout_ensemble}
%      \end{subfigure}
%      \caption{We present the evolution of hypervolume and the Pareto front for both the airfoil and the printer color gamut problems, utilizing LBN-MOBO with DE and MC dropout as the surrogate models. Figure (a) shows the hypervolume expansion of the Airfoil problem, and (b) represents the Pareto front calculated using each surrogate. Similarly, (c) depicts the hypervolume expansion of the printer color gamut problem, and (d) displays the gamut actually discovered by LBN-MOBO using both surrogates.}
%      \label{fig:airfoil_experiment_dropout_ensemble}
% \end{figure}
\begin{figure}
\centering
\subfigure[]{\label{fig:Progress_hv_airfoil_dropout_ensemble}\includegraphics[width=0.23\textwidth]{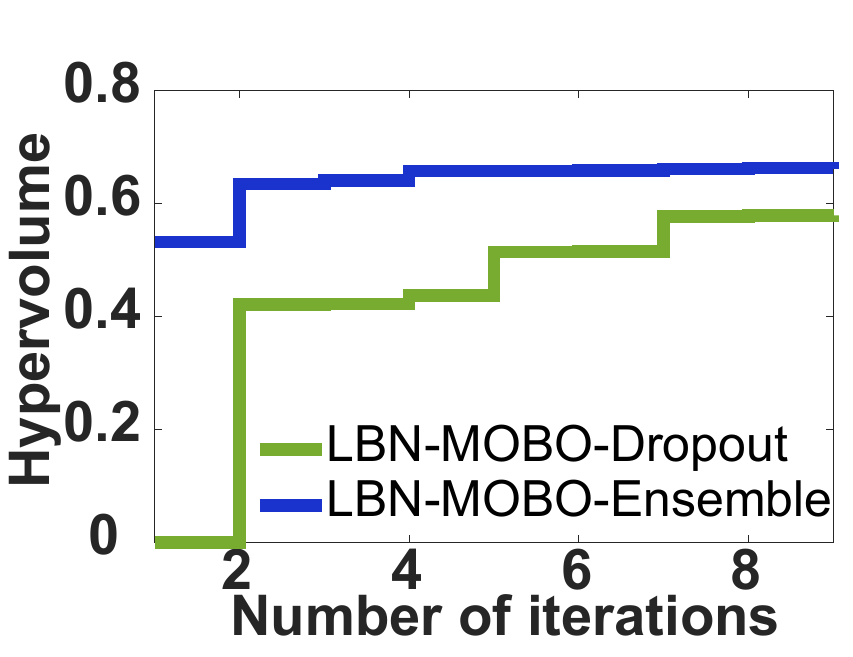}}
\subfigure[]{\label{fig:comparison_airfoil_dropout_ensemble}\includegraphics[width=0.23\textwidth]{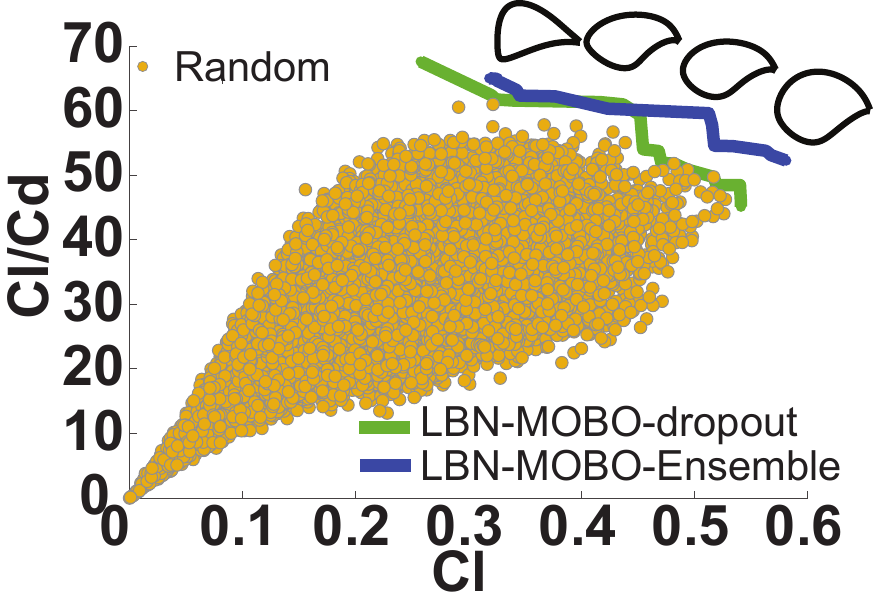}}
\subfigure[]{\label{fig:Progress_hv_AB_dropout_ensemble}\includegraphics[width=0.23\textwidth]{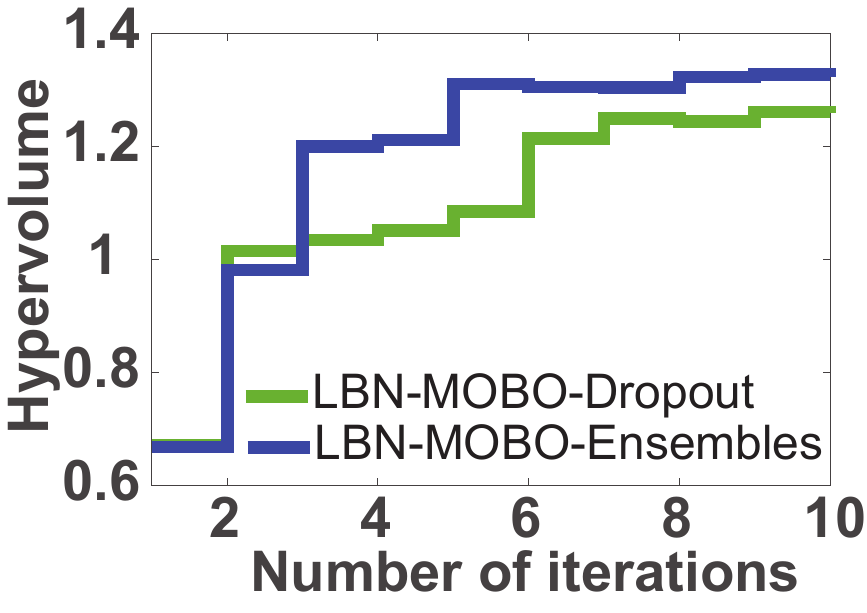}}
\subfigure[]{\label{fig:44_ink_gamut_dropout_ensemble}\includegraphics[width=0.23\textwidth]{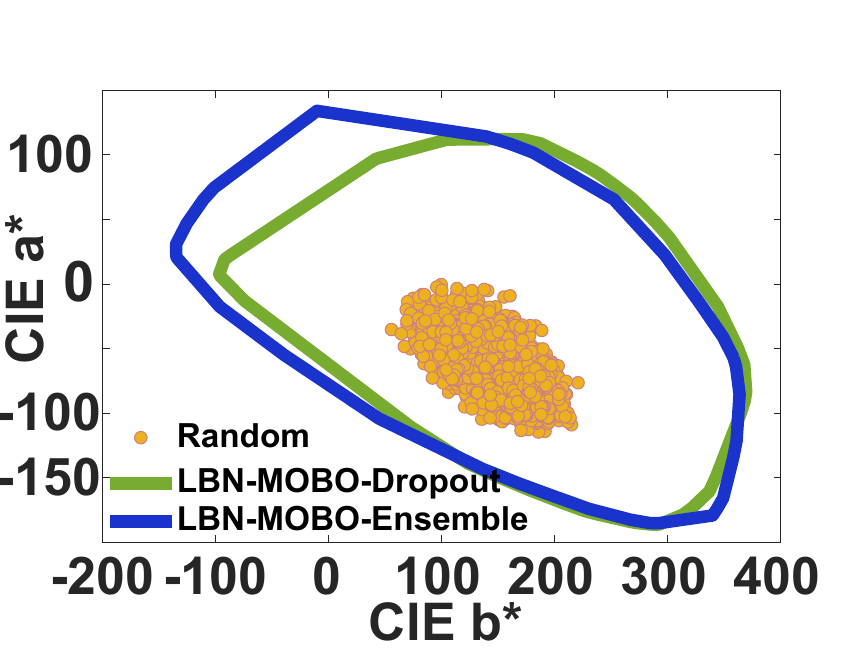}}
\caption{We present the evolution of hypervolume and the Pareto front for both the airfoil and the printer color gamut problems, utilizing LBN-MOBO with DE and MC dropout as the surrogate models. Figure (a) shows the hypervolume expansion of the Airfoil problem, and (b) represents the Pareto front calculated using each surrogate. Similarly, (c) depicts the hypervolume expansion of the printer color gamut problem, and (d) displays the gamut actually discovered by LBN-MOBO using both surrogates.}
\label{fig:airfoil_experiment_dropout_ensemble}
\end{figure}

In this study, we delve into the practical capabilities of LBN-MOBO by applying it to two complex real-world problems: airfoil and printer's color gamut, employing larger batch sizes of 15,000 and 20,000, respectively. The primary focus of this section is a comparison between LBN-MOBO with MC dropout and LBN-MOBO with DE. Both variations exhibit commendable performance in addressing the complexities of the selected problems, with DE showing a better performance. A comprehensive evaluation of additional methods on these problems is presented in Section~\ref{sec:complementary_eval_printer} and\ref{sec:complementary_eval_airfoil} of Appendix.
There we also establish the superiority of our method over a few other algorithms that can operate in this large batch regime. 

%%%%% Airfoil%%%%%%
Figure~\ref{fig:airfoil_experiment_dropout_ensemble} illustrates the analysis for the airfoil problem, showcasing the performance of LBN-MOBO with MC dropout and with DE. This experiment introduces the complex problem of mapping the airfoil shapes to intricate aerodynamic properties. Both variations of LBN-MOBO runs with batch size equal 15,000.
The candidate samples of each optimization iteration is simulated by the high-fidelity CFD solver OpenFOAM \cite{OpenFOAM}.
Remarkably, as depicted in Figure~\ref{fig:Progress_hv_airfoil_dropout_ensemble}, both LBN-MOBO variations are capable of handling this huge batch of data and uncover superior Pareto fronts in a limited number of iterations. 
We can observe the superior performance of LBN-MOBO utilizing DE due to its higher-quality epistemic uncertainty.
%
% For an extensive evaluation using advanced BO methods as well as multi-objective stochastic optimizations see Section~\ref{sec:complementary_eval_airfoil} of Appendix.

%%%%%% 44 ink Printer %%%%%%

% \begin{figure}
%      \centering
%      \begin{subfigure}[b]{0.45\textwidth}
%          \centering

%                 \includegraphics[width=\textwidth]{figs/Progress_hv_AB_dropout_ensemble.pdf}
%        \caption{Hypervolume evolution}
%          \label{fig:Progress_hv_AB_dropout_ensemble}
%      \end{subfigure}
%      \hfill
%      \begin{subfigure}[b]{0.45\textwidth}
%          \centering
%        \includegraphics[width=\textwidth]{figs/Gamut_AB_44ink_dropoit_ensemble.pdf}
%     \caption{Piecewise linear color gamut at iteration 10}
%     \label{fig:44_ink_gamut_dropout_ensemble}
%      \end{subfigure}
%      \hfill
%         \caption{The hypervolume evolution and color gamut of the printer calculated by different methods.}
%         \label{fig:hv_AB_dropout_ensemble}
% \end{figure}

For the printer's color gamut exploration, LBN-MOBO initiates with 10,000 samples, with each subsequent iteration managing a batch size of 20,000 samples. The high dimensionality of this design space (44) presents a formidable challenge, rendering it an intriguing test case. The performance space for this experiment is represented by the 2-dimensional a*b* color space.
Figure \ref{fig:Progress_hv_AB_dropout_ensemble} vividly demonstrates the large increase in the hypervolume of the color gamut achieved by LBN-MOBO. The final gamut estimation for both variations of LBN-MOBO after 10 iterations is illustrated in Figure~\ref{fig:44_ink_gamut_dropout_ensemble}, revealing the effectiveness of our method in estimating a significantly large gamut in a small number of iterations.
\subsection{The impact of epistemic uncertainty on the performance of LBN-MOBO} \label{sec:uncertaimty-effect}
% %
\begin{figure}
\centering
\subfigure[Airfoil]{\label{fig:LBN_Airfoil_uncertainty_ablation}\includegraphics[width=0.35\textwidth]{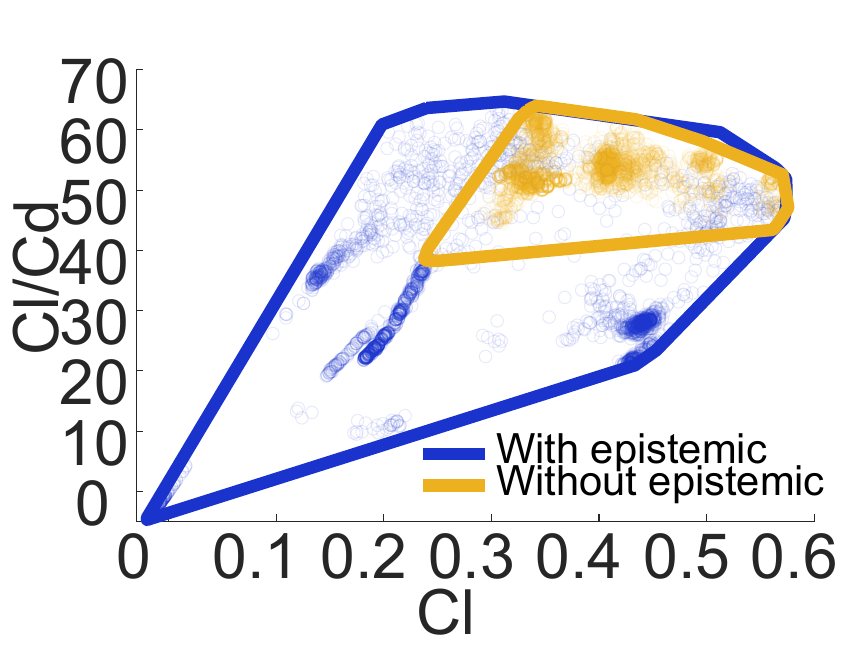}}
\subfigure[Color gamut]{\label{fig:abelation_AB}\includegraphics[width=0.35\textwidth]{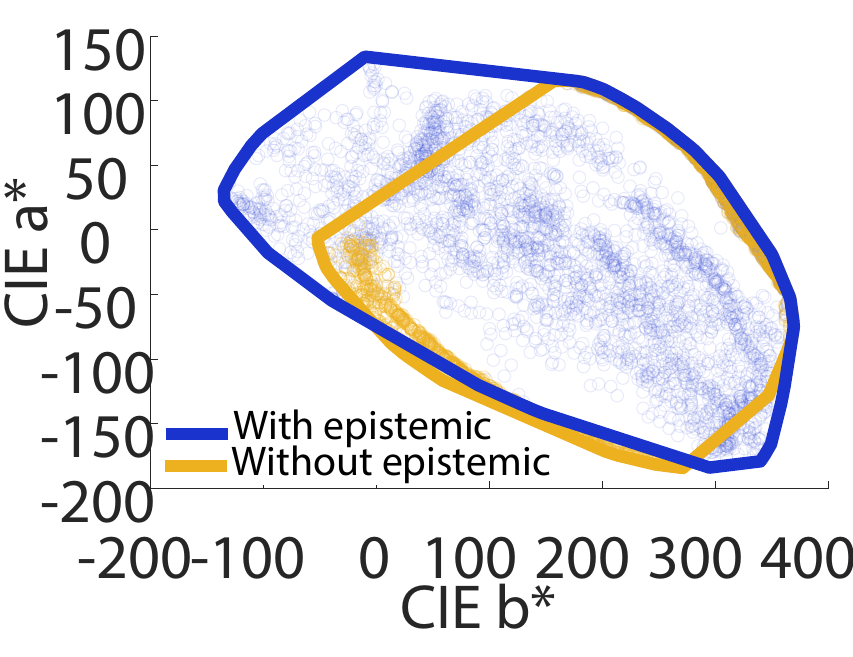}}
\caption{Ablation studies on the effect of epistemic uncertainty in our $2\textit{M}D$ acquisition function, using our real-world problems.}
\label{fig:Aifoil_test_case}
\end{figure}
One of the key factors enhancing the performance of LBN-MOBO is its use of uncertainty to effectively explore under-represented parts of the design space. 
We investigate the impact of uncertainty on the computation of the Pareto front for both airfoil design and color gamut exploration.
Both experimental setups mirror the conditions described in Section~\ref{sec:real_world}, except that they exclude uncertainty information. The candidate distribution from iteration 4 to 8 is illustrated in Figures \ref{fig:LBN_Airfoil_uncertainty_ablation} and ~\ref{fig:abelation_AB}.
For a clearer depiction of the samples' spatial distribution, we have illustrated their convex hull. Note that in the absence of uncertainty, the candidates have a tendency to cluster within particular areas. This clustering leads to diminished diversity and, as a consequence, a reduction in the capacity for exploration (as represented by the yellow samples).
Conversely, when uncertainty is incorporated into the process, we observe an increase in the diversity of the candidates and consequently, a broader Pareto front is discovered (represented by blue samples).
Furthermore, uncertainty guides the candidates to progressively bridge the information gap in the surrogate models, making them increasingly similar to the NFP. This factor further enhances the quality of the Pareto front retrieved through the LBN-MOBO process.

We also observe that when uncertainty is excluded from the process, the budget for surrogate Pareto front optimization is concentrated solely on performance dimensions. This concentration may occasionally lead to a slight local enhancement in optimization, as illustrated in the bottom-left quarter of Figure~\ref{fig:abelation_AB}.

\subsection{LBN-MOBO in the presence of severe noise}
\label{sec:LBN_MOBO_noise}

\begin{figure}
\centering
\subfigure[]{\label{fig:acquisition_plot_without_uncertainty}\includegraphics[width=0.30\textwidth]{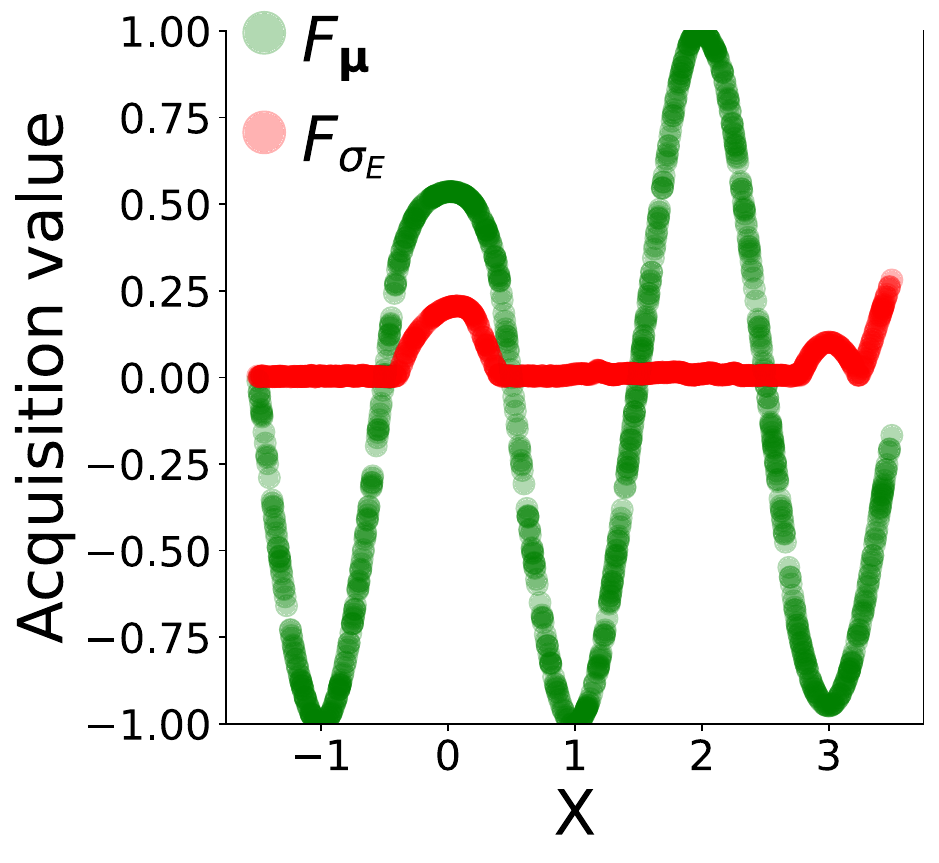}}
% \subfigure[]{\label{fig:acquisition_plot_with_uncertainty_w_1}\includegraphics[width=0.23\textwidth]{figs/acquisition_plot_with_uncertainty_w_1.pdf}}
\subfigure[]{\label{fig:acquisition_plot_with_uncertainty_w_7}\includegraphics[width=0.30\textwidth]{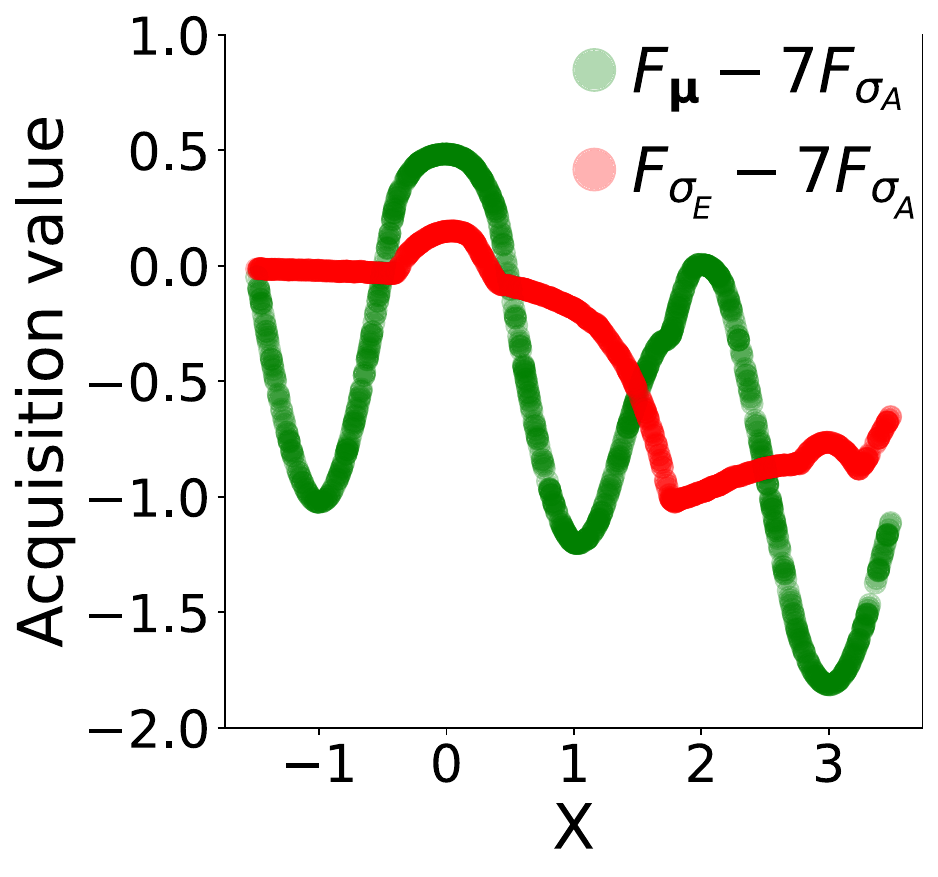}}
\subfigure[]{\label{fig:uncertainty_plot_toy}\includegraphics[width=0.30\textwidth]{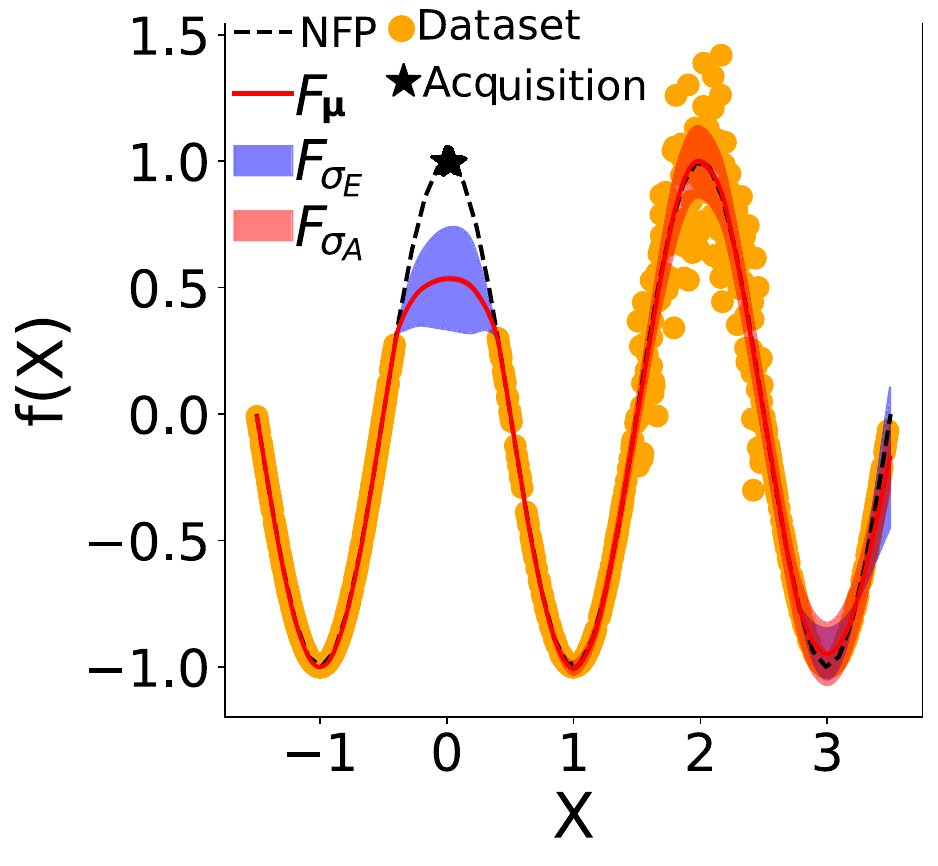}}
\caption{Modified LBN-MOBO performance in a toy problem with severe aleatoric noise using different weights for $\mathbb{F}_{\sigma A}(\mathbf{x})$. Once the importance of $\mathbb{F}_{\sigma A}(\mathbf{x})$ is adjusted correctly LBN-MOBO completely avoids regions with irreducible noise.}
\label{fig:Toy_noisy}
\end{figure}
As explained earlier modeling aleatoric uncertainty via DE in real-world, high-dimensional problems can be challenging and lead to unwanted behaviour of predicted aleatoric uncertainty $\mathbb{F}_{\sigma A}(\mathbf{x})$.
In this section we curate two examples to show that in case of having a good estimation of aleatoric uncertainty $\mathbb{F}_{\sigma A}(\mathbf{x})$, we can improve LBN-MOBO to become more robust against severe irreducible noise.
We begin with a straightforward toy example to elucidate the concept, followed by a real-world printer gamut experiment, which has been simplified to facilitate its aleatoric uncertainty estimation via DE. 

The toy example is defined as:
\small
\[
f(X) = 
\begin{cases} 
\begin{aligned}
    &\cos(\pi X) & \text{if } X \in [-1.5, -0.4]\cup [0.4, 1.5] \cup [2.5, 3.5]\\
\end{aligned} \\
\begin{aligned}
    &\cos(\pi X) + \mathcal{N}(0, 0.2) & \text{if } X \in [1.5, 2.5].
\end{aligned}
\end{cases}
\]
In this problem we have two equal local maximums, one of which is not sampled to introduce epistemic uncertainty and the other is contaminated with the Gaussian noise.
Our objective is to show the benefits of $\mathbb{F}_{\sigma A}(\mathbf{x})$ in avoiding regions with severe noise.
To this end, we augment all 2\textit{M}D objectives in our acquisition function with their corresponding aleatoric uncertainty values:
% \small
% \begin{equation} \label{eq:acquisition_aleatoric}
% \begin{aligned}
%      A\text{\tiny F} := \ \arg\max_{\mathbf{x}} &\left\{ \mathbb{F}_{\mathbf{\mu}}^{m}(\mathbf{x}) - \alpha \mathbb{F}^{m}_{\sigma A}(\mathbf{x}) \oplus \  \mathbb{F}_{\sigma E}^{m}(\mathbf{x})- \beta \mathbb{F}^{m}_{\sigma A}(\mathbf{x})\right\} \\
%      & m \in \left[ 1, M \right] \\
% \end{aligned}
% \end{equation}
\small
\begin{subequations} \label{eq:acquisition_aleatoric}
\small
\begin{gather}
 \Tilde{\mathbb{F}}(\mathbf{x}) = \mathbb{F}_{\mathbf{\mu}}^{m}(\mathbf{x})  - \alpha \mathbb{F}^{m}_{\sigma A}(\mathbf{x}) \oplus \  \mathbb{F}_{\sigma E}^{m}(\mathbf{x}) - \beta \mathbb{F}^{m}_{\sigma A}(\mathbf{x}), \ m \in \left[ 1, M \right]\\
     % \text{ParetoFront} := \ \arg\max_{\mathbf{x}} \Tilde{\mathbb{F}}(\mathbf{x}), \ \Tilde{\mathbb{F}}(\mathbf{x}) \in \mathcal{R}^{2M}\\
     A\text{\tiny F} (\Tilde{\mathbb{F}}(\mathbf{x}), S) := \text{ParetoFront}(\Tilde{\mathbb{F}}(\mathbf{x}), S)
     \end{gather}
\end{subequations}
where $\alpha$ and $\beta$ are the weights to balance the importance of aleatoric uncertainty during the optimization.
%

% \alireza{To this end, we augment all 2\textit{M}D objectives with their corresponding aleatoric uncertainty values:
% $$\Tilde{\mathbb{F}}(\mathbf{x}) 
% = \big(\mathbb{F}_{\mathbf{\mu}}^m(\mathbf{x})- \alpha \mathbb{F}^{m}_{\sigma A}(\mathbf{x}), \mathbb{F}_{\sigma E}^m(\mathbf{x}) - \beta\mathbb{F}^{m}_{\sigma A}(\mathbf{x})| ~m \in \left[ 1, M \right]\big),$$
% where $\alpha$ and $\beta$ are the weights to balance the importance of aleatoric uncertainty during the optimization. Our acquisition function outputs the set of Pareto-efficient solutions $A\text{\tiny F} (\Tilde{\mathbb{F}}(\mathbf{x}), S)$ that maximize $\Tilde{\mathbb{F}}(\mathbf{x})$.
% }
In Figure \ref{fig:Toy_noisy} we have compared one iteration of our algorithm with 3 different settings of $\alpha$ and $\beta$.
Figure \ref{fig:uncertainty_plot_toy} demonstrates our fitted model along with its estimated uncertainties. 
Since this is a 1D problem the acquisition function must find the Pareto front of a two dimensional optimization, corresponding to prediction $\mathbb{F}_{\mu}(\mathbf{x})$ and epistemic uncertainty $\mathbb{F}_{\sigma E}(\mathbf{x})$. 
Figure \ref{fig:acquisition_plot_without_uncertainty} shows these two optimization landscapes. If we run the default method without considering the aleatoric uncertainty we see that $\mathbb{F}_{\sigma E}(\mathbf{x})$ has a local maxima around $x=0$ as expected.
However the predicted local maxima is around $x = 2$ which is the unwanted noisy local maxima.
Running one iteration of LBN-MOBO on this problems with batch size 1000 leads to over 93\% of acquisition samples being chosen from the noisy region.
%

% \vahid{Running our method with aleatoric uncertainty} by setting $\alpha = 1$ and $\beta = 1$ in Equation \ref{eq:acquisition_aleatoric}, in Figure \ref{fig:acquisition_plot_with_uncertainty_w_1}, the shape of the loss landscape changes in favor of the local maxima at $x = 0$ by penalizing the noisy region. However, this change is not sufficient to completely avoid the noisy region.
%
In the next attempt, we set $\alpha$ and $\beta$ to $7$, leading to the loss landscapes presented in Figure \ref{fig:acquisition_plot_with_uncertainty_w_7}, which clearly favor the correct local maxima at $x=0$ for both objectives. The acquisition result of running LBN-MOBO for one iteration is presented in Figure \ref{fig:uncertainty_plot_toy} as black stars. Here, the algorithm has completely avoided the noisy maxima, with 100\% of its samples at $x = 0$.

\begin{figure}
\centering
\subfigure[]{\label{fig:AB_no_aleatoric}\includegraphics[height=0.25\textwidth]{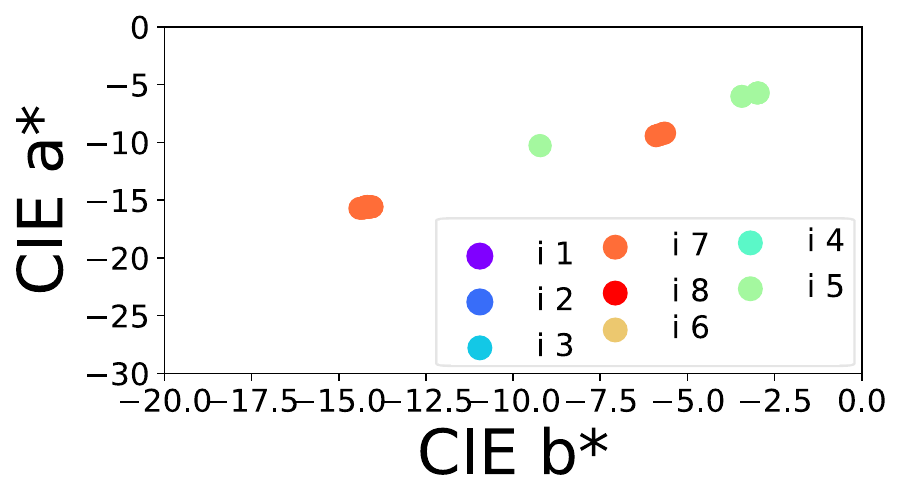}}
\subfigure[]{\label{fig:AB_aleatoric}\includegraphics[height=0.25\textwidth]{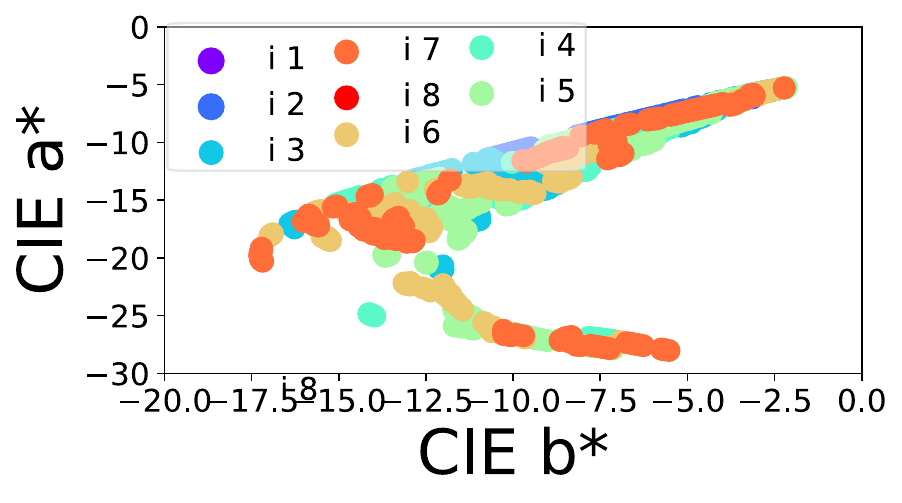}}
\subfigure[]{\label{fig:AB_aleatoric_w_10}\includegraphics[height=0.25\textwidth]{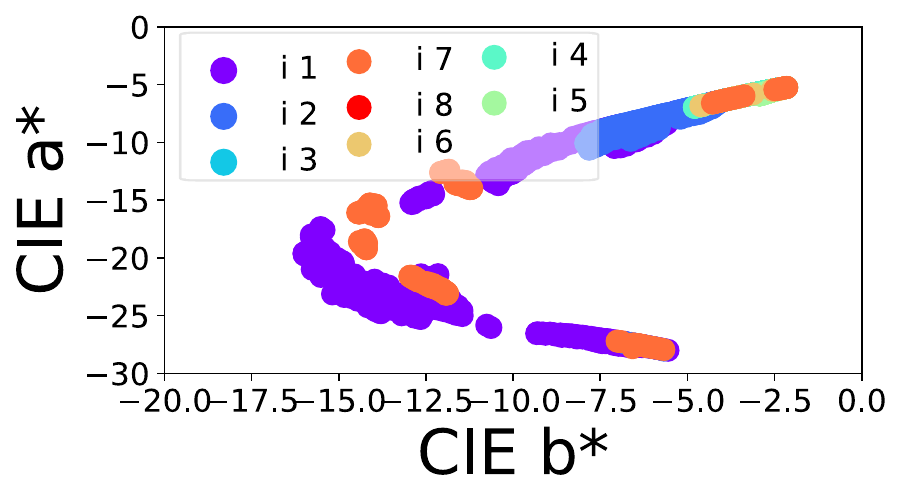}}
\subfigure[]{\label{fig:HV_uncertainty_ablation}\includegraphics[height=0.30\textwidth]{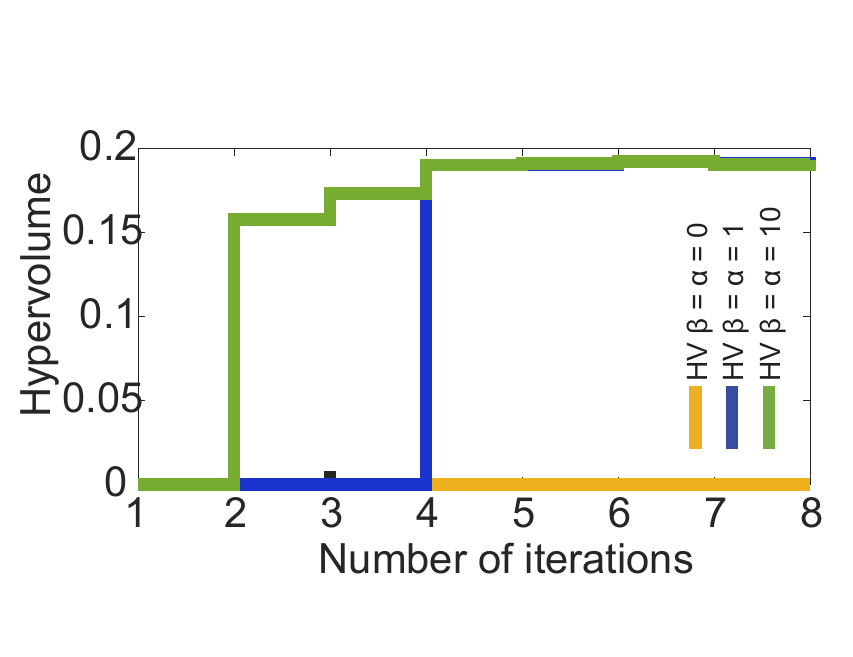}}
\subfigure[]{\label{fig:CIEa_noisy}\includegraphics[width=0.40\textwidth]{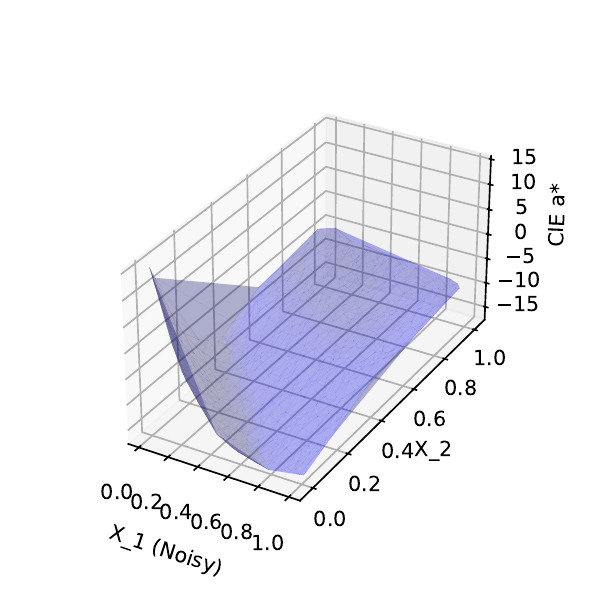}}
\subfigure[]{\label{fig:CIEb_noisy}\includegraphics[width=0.40\textwidth]{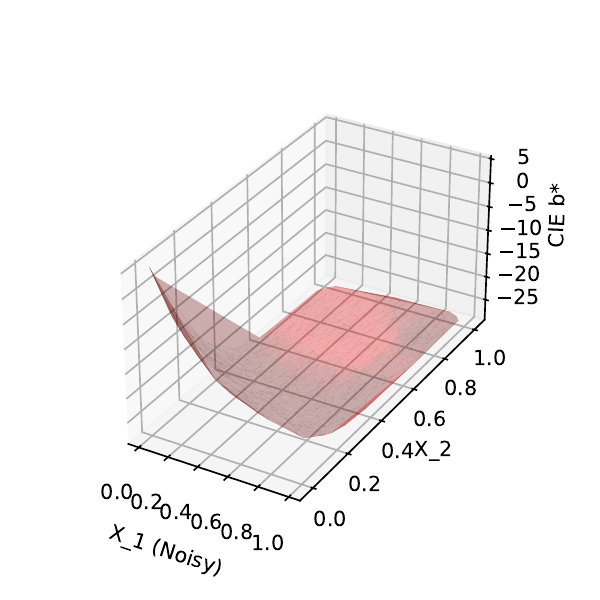}}
\caption{On this simplified printer's gamut problem, we fix 6 out of 8 channels and inject noise to the Light Cyan channel when it deposits inks with below 20\% intensity. We show that by considering aleatoric uncertainty we can make LBN-MOBO robust against this noise and still recover the correct color gamut.}
\label{fig:AB_printer_noise}
\end{figure}

Figure \ref{fig:AB_printer_noise} presents the real problem of the printer's color gamut, as discussed in Section \ref{sec:evaluation}.
To increase complexity, we added Gaussian noise (\(\mathcal{N}(0, 0.02)\)) to channel 4 (Light Cyan) whenever the ink intensity drops below 20\%.
Moreover, since capturing aleatoric noise in an 8-dimensional input problem via DE is challenging, we fix the values of 6 channels and vary only two.
The challenge in this problem is that the maxima lie entirely in noisy regions (Figures \ref{fig:CIEa_noisy} and \ref{fig:CIEb_noisy}). Consequently, discarding aleatoric uncertainty results in consistent sampling of these areas.
In this experiment, we run LBN-MOBO with a batch size of 1000 for 8 iterations, testing three different weights for aleatoric uncertainty.
Figure \ref{fig:AB_no_aleatoric} does not account for $\mathbb{F}_{\sigma A}(\mathbf{x})$, resulting in only 46 samples from noise-free regions after 8 iterations. Consequently, as indicated by the hypervolumes, it fails to compute a reasonable Pareto front.
After setting $\alpha$ and $\beta$ to $1$, LBN-MOBO recovers 1168 valid samples and calculates the Pareto front after four iterations. The valid samples and their iterations are shown in Figure \ref{fig:AB_aleatoric}.
Increasing $\alpha$ and $\beta$ to $10$ allows LBN-MOBO to generate 4982 samples and recover the Pareto front by iteration 2. As shown in Figure \ref{fig:AB_aleatoric_w_10}, after the second iteration, all samples focus on the local maxima.
These experiments clearly demonstrate LBN-MOBO's ability to become robust against irreducible aleatoric noise, provided that the surrogate model reliably captures it.
Reliable computing (and separation) of aleatoric and epistemic uncertainties for complex problems (large training data and multiple output dimensions) is a formidable challenge but lies outside the scope of this work.

% \section{Discussion} \label{sec:discussion}
% \vahid{Discussion/Limitation }\\
% \navid{limitation: a way to balance exploration and exploitation}\\
% \navid{Constraint handling}
% \navid{Noise handling}\\
% \navid{Active learning}\\
\section{Conclusion}
LBN-MOBO emerges as a potent optimizer for problems where an increase in the batch size does not significantly inflate simulation or experimentation costs, but iterations are expensive.
Notably, LBN-MOBO not only retrieves a superior Pareto front but also enhances the surrogate model throughout the optimization process, making it closely resemble the NFP. This implies that, within the context of active learning, this methodology could be implemented: starting with a random dataset and incrementally training the network with missing data until it converges to the NFP.
Looking forward, there are a few key aspects of this method that warrant further exploration. First, the potential of LBN-MOBO in managing design constraints needs to be assessed. Second, an analysis of its performance in the presence of noisy data could be undertaken, and possibly, it could be extended to enhance its robustness against noise.
Finally, while our current acquisition function is tuning-free, it is intriguing to explore explicit methods that manipulate the balance between exploration and exploitation and see how this balance affects the overall performance of LBN-MOBO.

% \vspace{36pt}
\newpage
\bibliographystyle{plainnat}
\bibliography{main}

\newpage
\noindent\fbox{%
  \colorbox{lightgray}{%
    \parbox{\dimexpr\textwidth-2\fboxsep-2\fboxrule\relax}{%
      \vspace{0.5cm} % Adjust this value as needed
      Our code is available at: \url{https://github.com/an-on-ym-ous/lbn_mobo}
      \vspace{0.5cm} % Adjust this value as needed
    }%
  }%
}

\appendix
\startcontents[appendices]
\section*{Appendix}
\printcontents[appendices]{}{1}{}
\newpage
% \section{Supplementary Material}
\section{Complementary Related Works}\label{sec:comp_related}
In addition to the neural BO models discussed in Section \ref{related}, several advanced BO algorithms, as well as stochastic multi-objective optimizations, are adept at handling multi-objective Bayesian optimization for moderate batch sizes.
In this section we review the most relevant state of the art methods that can address our class of problems.
In Sections \ref{sec:complementary_experiments}, we assess these methods using synthetic ZDT and DTLZ test suits, Printer color gamut, and Airfoil problems. We observe their stagnation with increasing batch sizes, generally lacking scalability for batches exceeding 1000 samples. 
Ultimately, we demonstrate that in terms of Pareto front optimality, these methods are not comparable to our LBN-MOBO algorithm.

\subsection{Stochastic multi-objective optimizations} \label{sec:NDS}

\textbf{Non-dominated sorting genetic algorithm II (NSGA-II)} \cite{deb2002fast} is an exceptionally popular method for multi-objective optimization. It belongs to multi-objective evolutionary algorithms (MOEA), which have been applied to a variety of problems, from engineering \cite{schulz2018interactive} to finance \cite{subbu2005multiobjective}.  
Despite their widespread use, MOEA have certain limitations.
One major challenge is the computational cost of MOEAs, as they typically require a large number of function evaluations, and iterations to converge to a good solution \cite{konakovic2020diversity}.
This can make MOEAs impractical for problems with computationally expensive objective functions or high-dimensional design spaces.
Moreover, they are prone to trap in local minima. This can be particularly problematic for problems with multiple local optima or non-convex objective functions. 

\subsection{Advanced Bayesian multi-objective optimizations}
{Bayesian optimization (BO)} is adept at efficiently searching for the global optimum while minimizing the number of function evaluations \cite{jones1998efficient}. However, extending BO to multi-objective batch optimization is not straightforward.
\textbf{USeMO} \cite{belakaria2020uncertainty} is one of the state of the art extensions of BO that is capable of solving multi objective problems. It employs NSGA-II to identify the Pareto front on the surrogate and uses uncertainty information to select a subset of candidates for the next iteration. 
\textbf{TSEMO} \cite{bradford2018efficient} takes a different approach by using Thompson sampling and NSGA-II on Gaussian process (GP) surrogates to find the next batch of samples that maximize the hypervolume. However, these methods struggle to maintain diversity and fail to capture part of the final Pareto front\cite{konakovic2020diversity}. 
% \vahid{this is the effect, how about difference in method} AAAAA
%
%

\textbf{Diversity-guided multi-objective Bayesian optimization (DGEMO)} seeks to address this issue by dividing the performance and design spaces into diverse regions and striving to identify candidates in as diverse locations as possible while maintaining the performance \cite{konakovic2020diversity}. However, its computational time grows exponentially with the increase in batch size.

\textbf{Thompson Sampling (TS)} has been the subject of significant research, with key contributions aiming to enhance the scalability of Bayesian Optimization. \cite{hernandez2017parallel} demonstrated its scalability in the chemical space through parallel and distributed computing, effectively handling large parallel measurements in BO. \cite{deshwal2021mercer} further extended its potentials in combinatorial BO settings by incorporating Mercer features, opening up new possibilities in molecular optimization. \cite{vakili2021scalable} addressed scalability challenges by integrating sparse Gaussian process models with Thompson Sampling. They provide a theoretical and empirical analysis proving that this scalable TS shows much less computational complexity while maintaining its performance quality. Scalable TS introduces new possibilities especially in the combinatorial space of high-throughput molecular design problems.

\subsection{Bayesian optimization and Pareto front of prediction and uncertainty}

\citet{gupta2018exploiting} proposed a unique algorithm that employs two distinct acquisition techniques to generate candidates for subsequent iterations. The primary insight of their method is to effectively handle problems characterized by a wide variety of local extrema, ranging from minimal to substantial in number. Their approach integrates the Gaussian Process Upper Confidence Bound (GP-UCB) with an additional Pareto front. This Pareto front is derived from optimizing predictions and uncertainties as separate objectives.

However, unlike the LBN-MOBO approach, Gupta et al.'s algorithm does not specifically focus on large batch optimization. Furthermore, it doesn't leverage Bayesian Neural Networks (BNNs) as surrogate models, which are key in enhancing scalability to the levels necessary for addressing complex real-world problems. Another distinct aspect of their work is that it appears primarily geared towards problems with single objectives, rather than the multi-objective scenarios that LBN-MOBO is designed to tackle.

\section{Complementary Methods} \label{complemntary_methods}
\subsection{Batch Pareto regret analysis for Bayesian optimization}
\label{sec:regret}
Analyzing the efficiency of an iterative optimization method is crucial for understanding its performance, and one effective way to do this is by monitoring its regret over time. Specifically, in Bayesian optimization, regret analysis offers insights into how efficiently the method approximates the global optimum over iterations. 

\subsubsection{Single-objective batch regret analysis}
Traditionally, regret analysis has been focused on single-objective sequential optimizations. In these settings, the optimization process involves calculating one sample candidate at a time, and the regret is measured based on the deviation from the optimal solution at each step. 
Let \( x^* = \arg\max_{x \in X} f(x) \) be the optimal solution. The instantaneous regret for a single recommendation \( x_t \) is defined as \( r_t = f(x^*) - f(x_t) \).

\citet{bubeck2012regret, bubeck2009pure, contal2013parallel} extended this concept by formulating the regret analysis for batch optimizations in single-objective scenarios. For a batch \( X_t \) of size \( n\), for each \( x_{i}^{t} \in X_t \), the regret \( r_{t}^{i} \) is defined as \( f(x^*) - f(x_{i}^{t}) \). We want to analyze the performance of the total cumulative regret for the batch of candidates throughout multiple iterations: 
% \begin{subequations}
% \begin{equation}
% R_{t}^n = \sum_{i=0}^{n} r_{t}^i \quad \text{(Total regret in a batch containing $n$ samples for a single iteration $t$)}
% \end{equation}
% \begin{equation}
% R_{T}^n = \sum_{t=0}^{t\leq T} R_t^n \quad \text{(Cumulative regret for batches across multiple iterations)}
% \label{eq:time_sum_regret}
% \end{equation}
% \end{subequations}
\begin{subequations}
\begin{align}
R_{t}^n &= \sum_{i=1}^{n} r_{t}^i \notag \\
&\text{(Total regret in a batch} \notag \\
&\text{containing } n \text{ samples for a single} \notag \\
&\text{iteration } t)
\end{align}
\begin{align}
R_{T}^n &= \sum_{t=0}^{t\leq T} R_t^n \notag \\
&\text{(Cumulative regret for} \notag \\
&\text{batches across multiple} \notag \\
&\text{iterations)}
\label{eq:time_sum_regret}
\end{align}
\end{subequations}

If the regret increases sub-linearly, it suggests that the method will eventually converge to the optimum if given enough iterations \cite{gupta2018exploiting}. This is formalized as:
\begin{equation}
    \lim_{T \to \infty} \frac{R_T^n}{T} = 0,
\end{equation}
which implies convergence. Given this, the following relationship holds:
\begin{equation}
    \min_{t \leq T} \{r_t\} \leq \frac{R_T^n}{T},
\end{equation}
leading to
\begin{equation}
    \lim_{T \to \infty} \max_{i, t \leq T} \{f(x_{t}^i)\} = f(x^*).
\end{equation}

\subsubsection{Multi-objective single sample regret analysis}
Further developments by \citet{xu2023pareto} introduce the concept of Pareto regret for multi-objective optimizations in the context of multi-objective multi-armed bandits (MO-MAB). Their approach, however, is limited to single-candidate scenarios and does not extend to batch optimization. 

In MO-MAB, the player chooses from \( n \) arms, each providing a \( D \)-dimensional reward vector over a time horizon \( T \). In multi-objective Bayesian optimization the problem translates to $f(x) \in \mathcal{R}^D$.

The Pareto regret in this context relates to the efficiency of choices made in this multi-objective space. Pareto optimality $P_F$, is defined as a set of the best solutions $\{f(x^*) \in P_F, \text{ for } x^* \in P_S\}$ that dominate all other solutions.
Pareto set ($P_S$) is the set of best designs that form the Pareto optimal front.

The Pareto regret for a single candidate $x$ at time \( t \) is given by:
\begin{equation}
    R_T = \sum_{t=0}^{T} \text{Dist}(f(x_t), P_F),
\end{equation}

where \( \text{Dist}(f(x_t), P_F) \) is a distance measure between a candidate solution \( f(x_t) \) and the \textit{set} \( P_F \). This measure evaluates how far the chosen candidate is from the Pareto optimal front.

Assume that $f(x^*) \in P_F$  is one of the solutions on the Pareto optimal front, \citet{xu2023pareto} define \( \text{Dist}(f(x_t), P_F) \) as:

% \begin{align}
% \text{Dist}(f(x_t), P_F) = \\ 
% & \min_{\epsilon \geq 0}\{\epsilon: f(x_t) +\epsilon \mathbf{1} || f(x*) \text{ for every } f(x^*) \in P_F \}.
% \end{align}
\begin{align}
\text{Dist}(f(x_t), P_F) &= \min_{\epsilon \geq 0} \big\{\epsilon: f(x_t) + \epsilon \mathbf{1} || f(x^*) \notag \\
&\qquad \forall f(x^*) \in P_F \big\}.
\end{align}
where  $\mathbf{1} \in \mathcal{R}^D$ is a unit vector with all elements equal to 1. Here, $\epsilon$ is a positive number that we iteratively increase until the new vector $f(x_t) +\epsilon \mathbf{1}$ in at least one dimension becomes larger than any of the solutions from Pareto optimal set. Finally, $a||b$ means that some dimensions of vector $a$ are larger than $b$ and some are smaller and as a result none of the vectors is dominating the other.

\subsubsection{Multi-objective, batch regret analysis}
Building on these foundational works, we propose a novel framework for regret analysis in \textbf{large batch}, \textbf{multi-objective} Bayesian optimization.
At every iteration we calculate the Pareto front of the proposed candidates at iteration $t$ as $P_f^t$ and calculate the average distance of all the elements of $P_f^t$ to the Pareto optimal front $P_F$:

\begin{equation}
R_{t}^n = \text{Dist}(P_f^t, P_F) =\frac{1}{n} \sum_{f(x_t) \in P_f^t} \text{Dist}(f(x_t), P_F), 
\label{eq:batch_MO_regret}
\end{equation}
 where $n$ is the number of solutions in current iteration of Pareto front $P_f^t$. The normalization helps take into account the fact that the number of solutions on Pareto front at each iteration is different.
 
Once we have the regret for a batch in a single iteration we can calculate the cumulative regret over multiple iterations using Equation~\ref{eq:time_sum_regret}.

In Sections \ref{sec:ZDT3}, \ref{sec:DTLZ1_DTLZ4} we applied this analysis on a range of benchmark problems and the results are presented in Figures  \ref{fig:ZDT3_regret}, \ref{fig:regrets_DTLZ4}, and \ref{fig:regrets_DTLZ1}. Moreover in Section \ref{sec:complementary_eval_airfoil} we analysed the regret on the real problem of Airfoil design and presented the results in Figure \ref{fig:regrets_airfoil}.
%%%%%%%%%%%%%%%%%%%%%
\subsection{MC dropout} \label{sec:MCdropout}

Incorporating MC dropout \cite{gal2016dropout} as a substitute for Deep Ensembles aims to leverage its inherent characteristics for assessing model uncertainty. MC dropout performs model uncertainty approximation by enabling dropout at the inference phase, generating diverse network predictions over multiple forward passes. These sub-networks will then ensemble in the same manner as our Deep Ensembles surrogate (Section \ref{sec:BNN}) to calculate the mean and uncertainty of the predictions.

\paragraph{MC Dropout as Surrogate Model}

For seamless integration with LBN-MOBO, we employ a Neural Network with dropout layers, designated as $f_{MC}$.
During the inference the dropout layers remain active, resulting in varied network structures for each forward pass. This stochasticity during inference results in a distribution of predictions for any given input, enabling the estimation of epistemic uncertainty.

A primary advantage of utilizing MC dropout lies in its capacity to compute uncertainties in parallel, akin to Deep Ensembles, hence maintaining the scalability of LBN-MOBO. After fitting $f_{MC}$ to the initial samples, the subsequent steps in LBN-MOBO remain consistent, with the updated surrogate model being utilized to compute the novel acquisition function.

\paragraph{Epistemic Uncertainty through MC Dropout}

MC dropout facilitates the calculation of epistemic uncertainty by observing the variance in predictions across multiple stochastic forward passes. Given a set of $T$ stochastic forward passes, the epistemic uncertainty for an input $\mathbf{x}$ can be computed as follows:

\begin{equation}
\mathbb{F}_{\mu_{MC}}(\mathbf{x}):= \frac{1}{T} \sum_{t} \mu_{t}(\mathbf{x}) \label{eq:MC_mu}
\end{equation}

\begin{equation}
\mathbb{F}_{\sigma_{MC}}(\mathbf{x}) = \frac{1}{T} \sum_{t} (\mu_{t}^{2}(\mathbf{x}) - \mathbb{F}_{\mu_{MC}}^{2}(\mathbf{x})). \label{eq:MC_sigma}
\end{equation}

\paragraph{Modified Acquisition Function with MC Dropout}

With MC dropout incorporated as the surrogate model, the novel acquisition function now utilizes the uncertainties and predictions obtained from $f_{MC}$. The modified acquisition function aims to balance the trade-off between exploiting regions of high predicted performance and exploring regions with high uncertainty, as determined by the MC dropout model.

\begin{equation} \label{eq:MC_acquisition}
A\text{\tiny F}_{\tiny MC} := \arg\max_{\mathbf{x}} \left\{ \mathbb{F}_{\mu_{MC}}^{m}(\mathbf{x}), \mathbb{F}_{\sigma_{MC}}^{m}(\mathbf{x})\right\} , m \in \left[ 1, M \right] .
\end{equation}

This integration of MC dropout with the novel acquisition function ensures that the benefits of epistemic uncertainty estimation are harnessed effectively while maintaining the scalability and parallelism inherent to the LBN-MOBO framework.

\subsection{Combination of other BNNs with 2\textit{M}D acquisition function}
\label{sec:BNNs_2MD}
\begin{figure*}
    \centering        
    \includegraphics[width=0.7\textwidth]{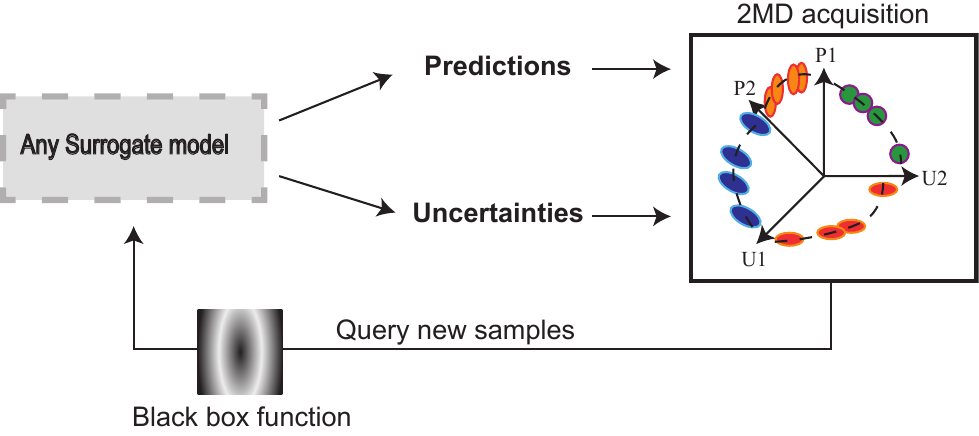}
    \caption{2\textit{M}D acquisition function can be paired with any surrogate model.}
    \label{fig:2MD_mechanism}
\end{figure*}
% %

One of the main advantages of 2\textit{M}D acquisition function is that it only queries a surrogate model that provides the prediction and uncertainty (without the need for gradient information). Figure \ref{fig:2MD_mechanism} demonstrates how 2\textit{M}D performs by iteratively using this queried information and evolves the solution until a practically good convergence. This property of 2MD acquisition allows it to be paired easily with any form of surrogate model.
% \subsection{\revised{Batch, Pareto regret analysis}}

% One of the best ways to analyze the efficiency of an iterative optimization method is to monitor its regret as it goes forward. If the increase of regret is sub-linear it means that if we continue the optimization indefinitely eventually we converge. 
% %
% Typically regret analysis is designed for single objective sequential optimizations in which at a time we calculate a single sample candidate.
% %
% XXX formulated the regret for a batch optimization for a single objective ...
% % 
% More xxx propose an approach for Pareto regret analysis for a multi-objective optimization however for a single batch.
% %
% Inspired by these ideas, here for the first time we present the regret analysis for batch, multi-objective optimization.
%

\section{Complementary Evaluation Details}\label{sec:complementary_experiments}

\subsection{ZDT problems} \label{sec:ZDT_problems}
The ZDT (Zitzler-Deb-Thiele) test suite \cite{zitzler2000comparison} is a set of benchmark problems commonly used for testing the performance of multi-objective optimization algorithms. The suite consists of six problems (ZDT1 through ZDT6), each having two objective functions. Here, we describe the formulations for ZDT1, ZDT2, and ZDT3.
\subsubsection{ZDT1}

ZDT1 is a 30-dimensional problem defined as follows:

Objective 1:
\[
f_1(x) = x_1
\]

Objective 2:
\[
f_2(x) = g(x) \left[ 1 - \sqrt{\frac{x_1}{g(x)}} \right]
\]

where:
\[
g(x) = 1 + \frac{9}{n-1} \sum_{i=2}^{n} x_i
\]
and \(x_i\) is in the range [0,1].

\subsubsection{ZDT2}
ZDT2, like ZDT1, is also a 30-dimensional problem and is defined as follows:

Objective 1:
\[
f_1(x) = x_1
\]

Objective 2:
\[
f_2(x) = g(x) \left[ 1 - \left(\frac{x_1}{g(x)}\right)^2 \right]
\]

where:
\[
g(x) = 1 + \frac{9}{n-1} \sum_{i=2}^{n} x_i
\]
and \(x_i\) is in the range [0,1].

\subsubsection{ZDT3}
ZDT3, a 30-dimensional problem, introduces a discontinuous Pareto front. The objectives are defined as:

Objective 1:
\[
f_1(x) = x_1
\]

Objective 2:
\[
f_2(x) = g(x) \left[ 1 - \sqrt{\frac{x_1}{g(x)}} - \frac{x_1}{g(x)} \sin(10\pi x_1) \right]
\]

where:
\[
g(x) = 1 + \frac{9}{n-1} \sum_{i=2}^{n} x_i
\]
and \(x_i\) is in the range [0,1].

\subsubsection{Benchmarking LBN-MOBO on ZDT3} \label{sec:ZDT3}

In this study, we highlight the superior performance of the LBN-MOBO compared to a set of state-of-the-art Multi-objective Bayesian optimizations, namely USeMO, DGEMO, TSEMO, and NSGA-II, on the ZDT3 test.
We demonstrate how LBN-MOBO adeptly manages large design spaces while maintaining a significantly shorter optimization time compared to its counterparts.
This investigation involves two ZDT3 problem configurations. The first experiment focuses on a 6-dimensional design space, while the second broadens this space to 30 dimensions, thereby increasing the complexity of the search space.
We maintain a fair comparison by limiting the batch size to 1000 samples for all algorithms despite the fact that LBN-MOBO inherently possesses the ability to handle much larger batches.
Using 1000 samples is an approximate limit of tractability for most of the competing methods.
Moreover throughout this experiment we use an equal number of iterations for all methods, except in cases where a method becomes intractable due to unmanageable computational load.

\begin{figure*}
    \centering
        
        \includegraphics[width=\textwidth]{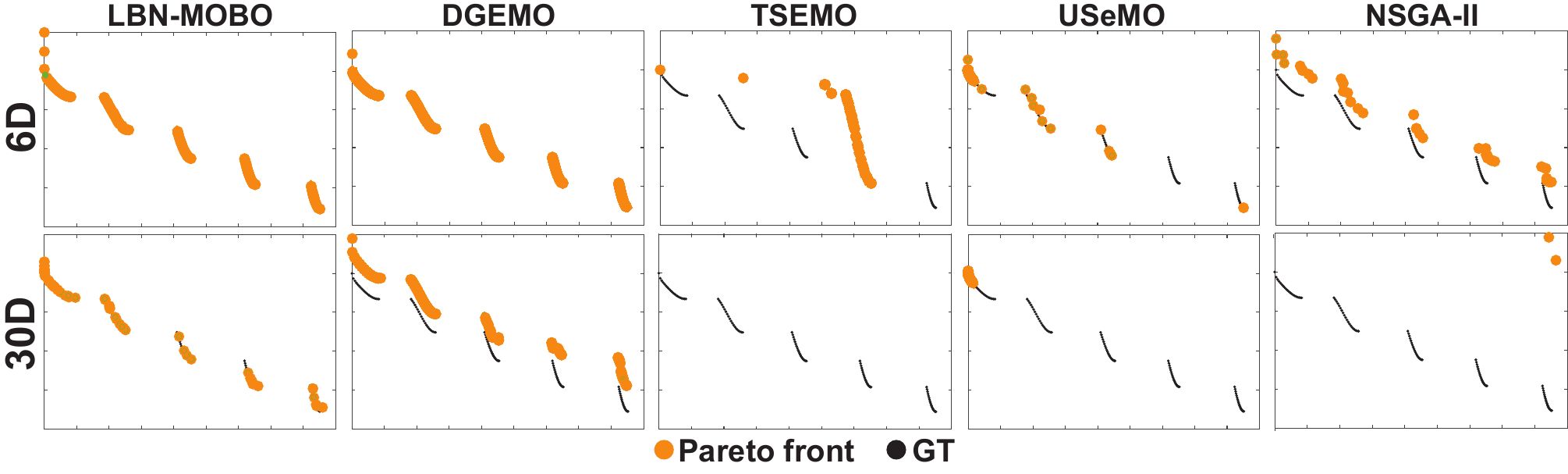}
    \caption{Obtained Pareto front by different methods on 6 dimensional and 30 dimensional ZDT3 problem. Batch size for all experiments is fixed at 1000.}
    \label{fig:zdt3_pareto}
\end{figure*}

Figure \ref{fig:zdt3_pareto} (top) shows the superior performance of LBN-MOBO and DGEMO in contrast with the other algorithms for the 6-dimensional ZDT3 problem. In this figure, the illustrated Pareto fronts are the final results of 10 optimization iterations. 
When confronted with the 30 dimensional problem (bottom row), the optimization methods must navigate a significantly larger space within the same sampling constraints. For this problem, we have shown the results of optimizations at iteration 5. 

\begin{figure*}
    \centering
        \includegraphics[width=\textwidth]{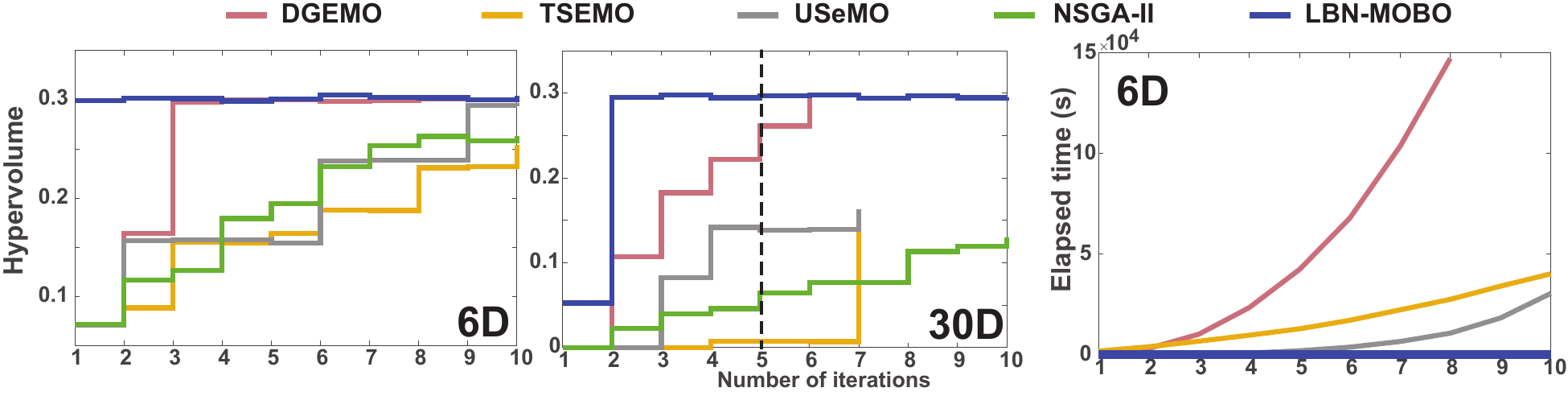}
    \caption{The left and middle plot represent the hypervolume of the 6 and 30 dimensional ZDT3 problem, respectively. The plot on the right shows the elapsed time for 6D problem for all methods (the NSGA-II plot is masked under LBN-MOBO).}
    \label{fig:hv_time_zdt3}
\end{figure*}

In Figure~\ref{fig:hv_time_zdt3} (left and middle), we further clarify the performance of different methods by showing the evolution of the hypervolume of the Pareto front at each iteration. 
For the 6D problem, LBN-MOBO manages to find the Pareto front after a single iteration. 
For the 30D problem, LBN-MOBO finds the Pareto front after two iterations and maintains its dominance over the other methods until the 6th iteration where DGEMO reaches it. 
We note that to approximate the hypervolume, we employed \cite{Simone2023} that uses random sampling. As a result, occasional minor fluctuations may arise.
It's worth mentioning that for the 30D problem we were unable to complete 10 iterations for UseMO, TSEMO, and DGEMO due to their exponential rise in computational time. 

Figure \ref{fig:hv_time_zdt3} (right) depicts the run-time of all methods for 6D ZDT3 for 10 iterations.
Even in this fairly straightforward problem, the computational time for UseMO, TSEMO, and particularly DGEMO surged dramatically. Conversely, the total computational time for NSGA2 was less than 14 seconds. 
The last and longest iteration of LBN-MOBO was 167s for training the ensemble models and 13.5s for acquisition.

\begin{figure}
\centering
\subfigure[ZDT3-6D]{\label{fig:regrets_ZDT3_6D}\includegraphics[width=85mm]{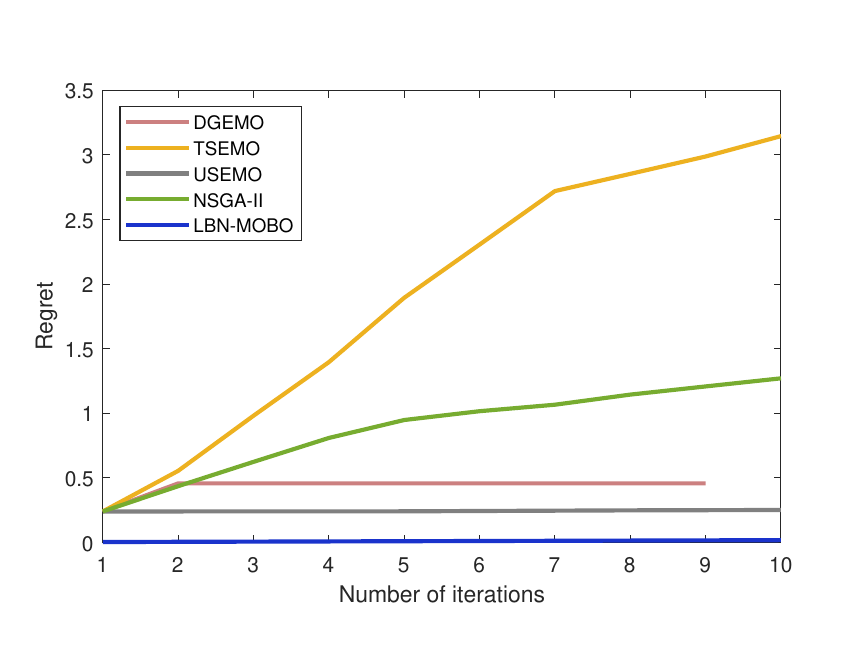}}
\subfigure[ZDT3-30D]{\label{fig:regrets_ZDT3_30D}\includegraphics[width=85mm]{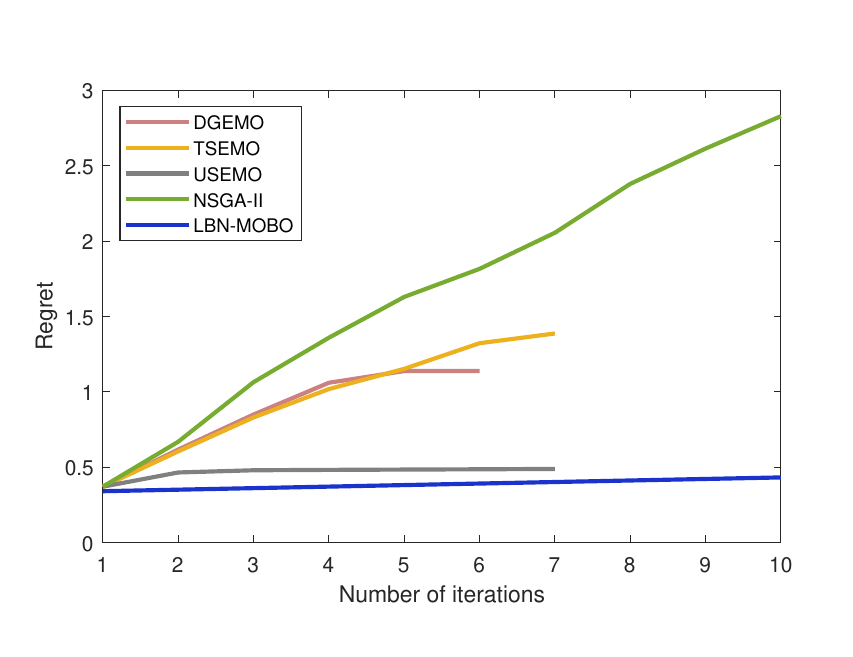}}
\caption{ZDT3 regret analysis for both 6D and 30D input setting.}
\label{fig:ZDT3_regret}
\end{figure}

{Additionally, Figure \ref{fig:ZDT3_regret} illustrates the cumulative regret associated with these optimizations, serving as a metric to quantify the divergence between the Pareto front of the optimal solutions at each iteration and the Pareto optimal front. This metric is particularly important in evaluating the proximity of the obtained solutions to the ideal outcomes.}

{The regret analysis results for both the 6D and 30D configurations underscore the superior performance of the LBN-MOBO approach in comparison to other alternatives. These findings highlight the effectiveness of LBN-MOBO in navigating the solution space towards the Pareto optimal front, as evidenced by the steady cumulative regret values across iterations.}
% %

%
%
% \begin{figure}
%      \centering
%      \begin{subfigure}[b]
%          \centering
%                 \includegraphics[width=85mm]{figs/regrets_ZDT3_6D.pdf}
%        \caption{ZDT3-6D}
%          \label{fig:regrets_ZDT3_6D}
%      \end{subfigure}
%      \hfill
%      \begin{subfigure}[b]
%          \centering
%        \includegraphics[width=85mm]{figs/regrets_ZDT3_30D.pdf}
%     \caption{ZDT3-30D}
%     \label{fig:regrets_ZDT3_30D}
%      \end{subfigure}
%      \hfill

%         \caption{ZDT3 regret analysis for both 6D and 30D input setting.}
%         \label{fig:ZDT3_regret}
% \end{figure}

\begin{figure*}
    \centering
        
        \includegraphics[width=\textwidth]{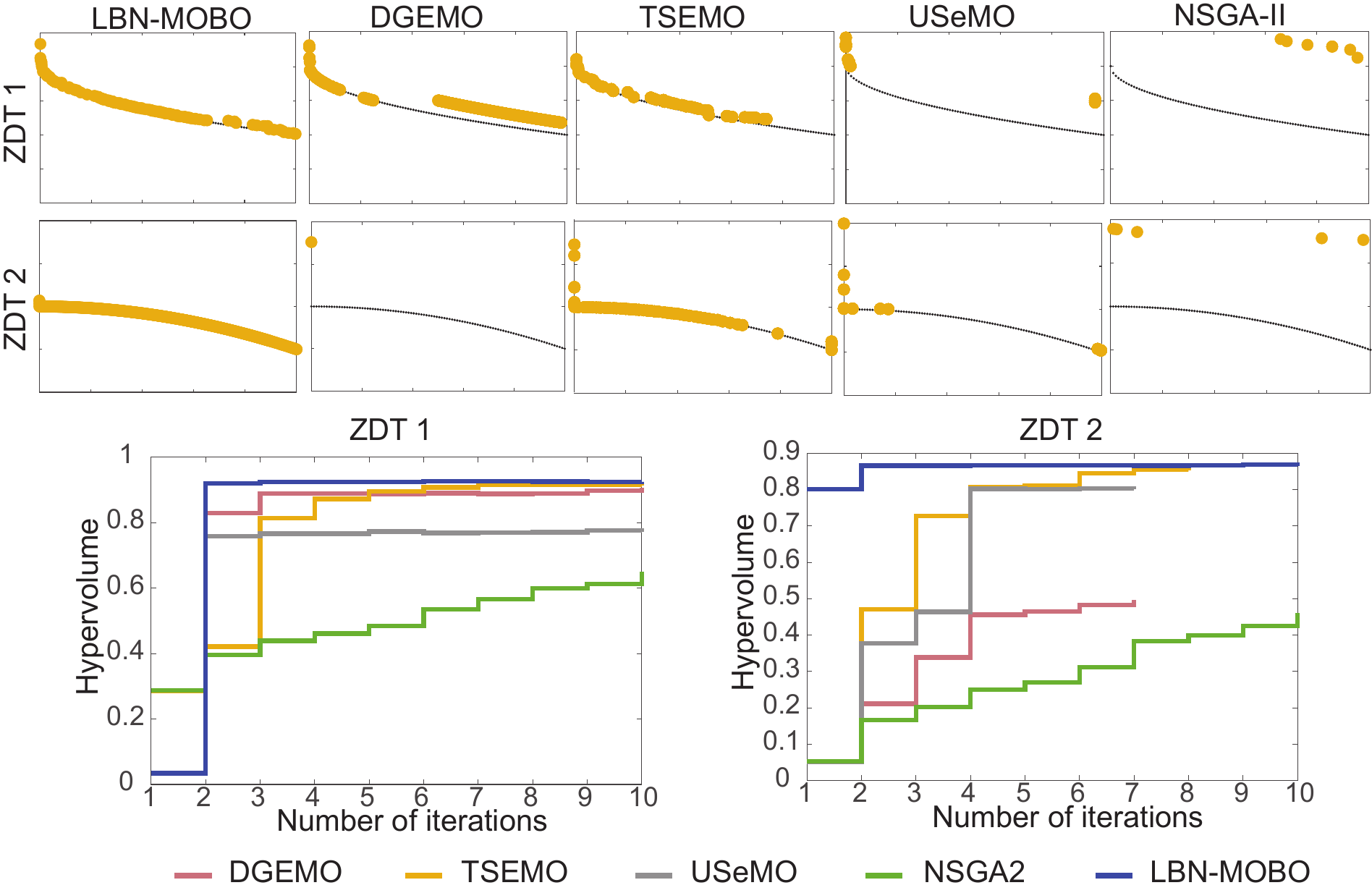}
    \caption{ZDT 1 and ZDT 2 in 30 dimensional setting. Note the immediate convergence of LBN-MOBO to Pareto front (bottom). The Pareto front is presented after 10 iterations of optimization except for USeMO (7 iterations), DGEMO (7 iterations), and TSEMO (8 iterations) in ZDT2 problem.} 
    \label{fig:ZDT1_2}
\end{figure*}
\subsubsection{Further evaluation using ZDT1 and ZDT2} \label{sec:ZDT1_ZDT2}
In this section, we showcase ZDT1 and ZDT2, two problems from the ZDT test suite \cite{zitzler2000comparison}.
Both problems involve conflicting objectives, with the only distinction that ZDT1 has a convex Pareto front, whereas ZDT2 has a non-convex one.
Both tests are conducted using their original 30-dimensional design space.
The problem setup configuration is entirely identical to that of the ZDT3 problem Section 4.2 in the paper.
As illustrated in Figure \ref{fig:ZDT1_2}, similar to the case of ZDT3, LBN-MOBO demonstrates its superiority in both the ZDT1 and ZDT2 problems.

The Pareto front in Figure \ref{fig:ZDT1_2} for the ZDT1 problem was obtained after 10 iterations for all the methods considered in the analysis. As for the ZDT2 problem, USeMO and DGEMO were able to run for 7 iterations, while TSEMO managed to run for 8 iterations before becoming intractable.
\subsection{DTLZ test suite with 3 dimensional output}
The DTLZ test suite is a popular benchmark set of test problems used for testing the performance of multi-objective optimization algorithms \cite{deb2005scalable}. The DTLZ test suite is characterized by scalable problems, meaning that the number of objectives and decision variables can be easily adjusted. This feature makes it particularly useful for assessing how well algorithms handle problems of varying complexity.

Below are the definitions of the DTLZ1, DTLZ4, and DTLZ5 problems, which we used to benchmark LBN-MOBO for 3D output.
\subsubsection{DTLZ1}
DTLZ1 is defined as:

Objective function \( f_m \) is given by:
\[
f_m(\mathbf{x}) = \frac{1}{2} \left(1 + g(\mathbf{x})\right) \prod_{i=1}^{m-1}(1 - x_i)
\]

\[
 g(\mathbf{x}) = 100 \left( |x| + \sum_{i=1}^{|x|} \left( x_i - 0.5 \right)^2 - \cos(20 \pi (x_i - 0.5)) \right)
\]

\[
0 \leq x_i \leq 1, \quad \text { for } i=1,2, \ldots, n
\]
where \( m \) is the number of objectives. Here, \( x \) denotes the decision variables vector.

\subsubsection{DTLZ4}
DTLZ4 is similar to DTLZ1, but with an additional parameter \( \alpha \) that controls the shape of the Pareto front. The objective functions are defined as:

\[
f_m(\mathbf{x}) = (1 + g(\mathbf{x})) \prod_{i=1}^{m-1} \cos\left(x_i^{\alpha} \frac{\pi}{2}\right)
\]

\[
0 \leq x_i \leq 1, \quad \text { for } i=1,2, \ldots, n
\]
where \( g(\mathbf{x}) \) is the same as in DTLZ1, \( m \) is the number of objectives, and \( \alpha \) is a user-defined parameter, typically set to 100, which controls the density of solutions.

\subsubsection{DTLZ5}
DTLZ5 is formulated to test the algorithm's ability to converge to a curved Pareto front and to maintain diversity among solutions. The DTLZ5 problem is defined as follows:

Objective functions:
% \begin{equation}
%     f_i(\mathbf{x}) = (1 + g(\mathbf{x}_M)) \cos\left(x_1 \frac{\pi}{2}\right) \ldots \cos\left(x_{i-1} \frac{\pi}{2}\right) \sin\left(x_{i} \frac{\pi}{2}\right), \quad \text { for } i=1,2, \ldots, m
% \end{equation}
\begin{align}
    f_i(\mathbf{x}) = & (1 + g(\mathbf{x}_M)) \cos\left(x_1 \frac{\pi}{2}\right) \times \notag \\
    & \times \ldots \times \cos\left(x_{i-1} \frac{\pi}{2}\right) \times \notag \\
    & \times \sin\left(x_{i} \frac{\pi}{2}\right), \quad \text{for } i=1,2, \ldots, m
\end{align}

where
\begin{itemize}
    \item \( \mathbf{x} = (x_1, x_2, \ldots, x_n) \) are the decision variables.
    \item \( m \) is the number of objectives.
    \item \( g(\mathbf{x}_M) \) is a function defined as: \( g(\mathbf{x}_M) = \sum_{x_i \in \mathbf{x}_M} (x_i - 0.5)^2 \), with \( \mathbf{x}_M \) being the subset of decision variables starting from the \( m \)th variable to the \( n \)th variable.
\end{itemize}
The constraints are:
\begin{equation}
    0 \leq x_i \leq 1, \quad \text { for } i=1,2, \ldots, n
\end{equation}

DTLZ5's primary challenge lies in its reduced dimensionality of the search space due to the use of trigonometric functions, which tend to align the solutions along a curve in the objective space. This problem is particularly useful for assessing the ability of an optimization algorithm to handle non-linear relationships between objectives and to generate a well-distributed set of solutions along a curved Pareto front.

\subsection{Benchmarking LBN-MOBO against other BNNs on DTLZ5 problem}
\label{sec:DTLZ5}
\begin{figure*}
    \centering
        \includegraphics[width=\textwidth]{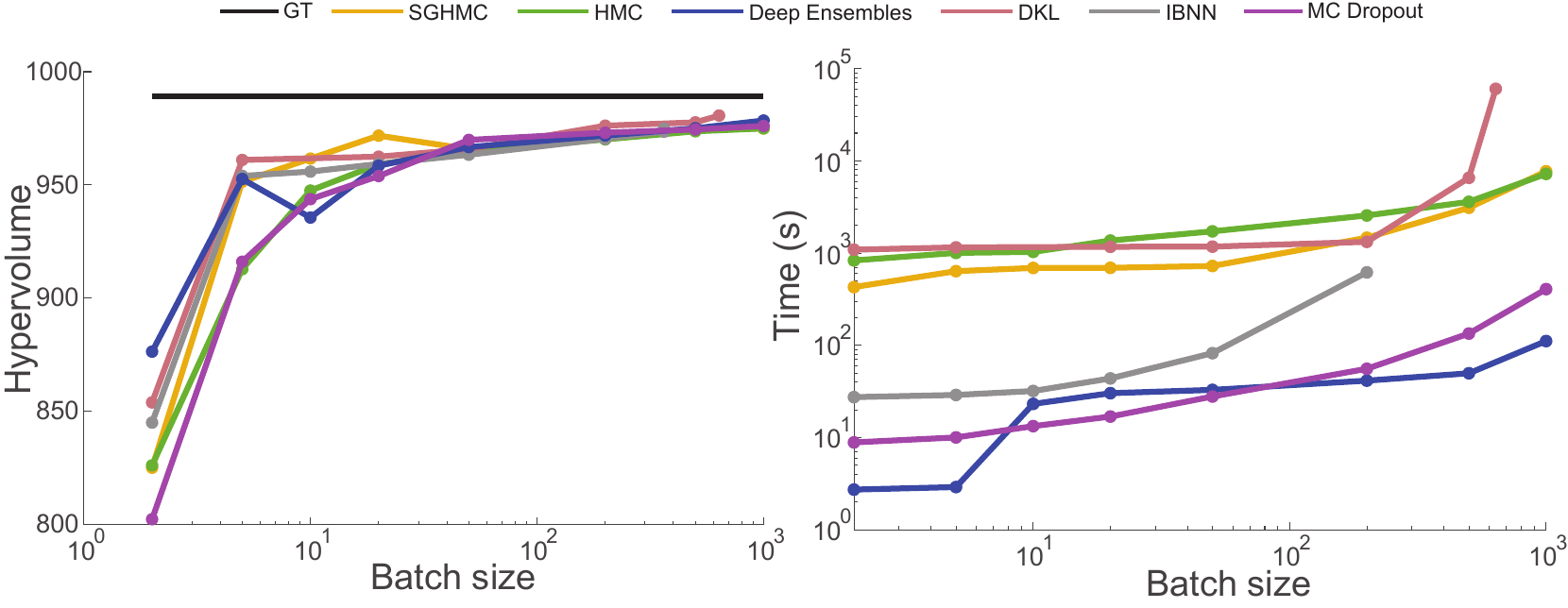}
    \caption{DTLZ5 benchmark optimization via 2\textit{M}D acquisition function and various neural BO surrogate models. The fitting time of each model was recorded to assess computational efficiency.}
            \label{fig:DTLZ5}
\end{figure*}

The DTLZ5 problem, with its extensive input and output dimensions, presents a formidable challenge in optimization. To showcase the robustness and scalability of our pipeline, we applied it to the DLTZ5 test, characterized by 20 input design parameters and 3 output dimensions. As illustrated in Figure \ref{fig:DTLZ5}, this problem displays behavior akin to what we observed in our earlier tests (see Figure \ref{fig:2MD_time_hv}). Although all the methods we tested show comparable performance, dropout and Deep Ensembles particularly distinguish themselves with their notably higher scalability.
The setup of this experiment is identical to the experiment in Section \ref{sec:selection}.

\subsection{Benchmarking LBN-MOBO on DTLZ1 and DTLZ4 using a variety of other optimization techniques}
\label{sec:DTLZ1_DTLZ4}
\begin{figure*}
    \centering
        \includegraphics[width=\textwidth]{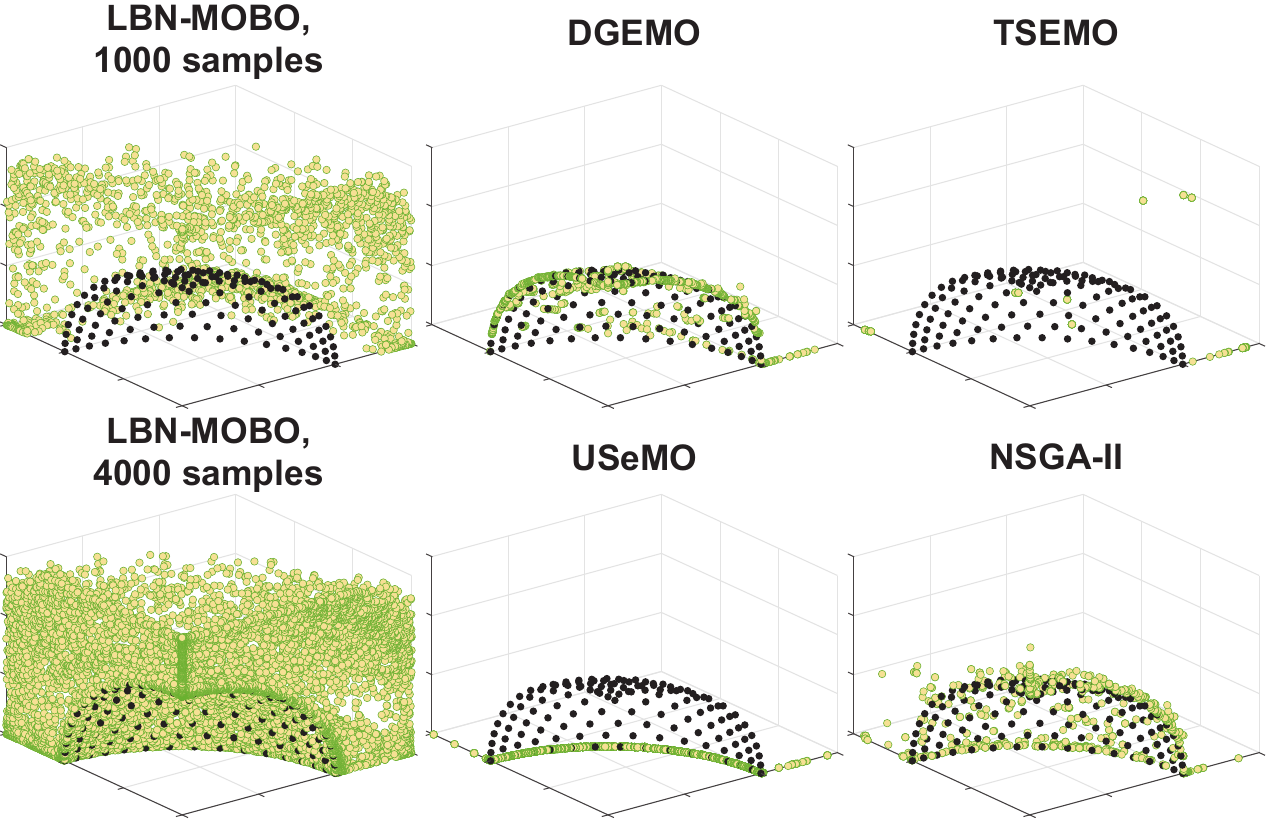}
    \caption{The Pareto front of the DLTZ4 problem with 6-D input and 3-D output.}
            \label{fig:DLTZ4}
\end{figure*}
%
% \begin{figure}
%      \centering
%      \begin{subfigure}
%                 \includegraphics[width=85mm]{figs/Time_DTLZ4.pdf}
%        \caption{Optimization time}
%          \label{fig:DLTZ4_time}
%      \end{subfigure}
%      \begin{subfigure}
%        \includegraphics[width=85mm]{figs/Progress_hv_DTLZ4.pdf}
%     \caption{Hypervolume expansion}
%     \label{fig:DLTZ4_hv}
%      \end{subfigure}
%           \begin{subfigure}
%        \includegraphics[width=85mm]{figs/regrets_DTLZ4.pdf}
%     \caption{Regret analysis}
%     \label{fig:regrets_DTLZ4}
%      \end{subfigure}
%         \caption{DTLZ4 experiment with 6 inputs and 3 outputs.}
%         \label{fig:DLTZ4_hv_time}
% \end{figure}
%
\begin{figure}
\centering     %%% not \center
\subfigure[Optimization time]{\label{fig:DLTZ4_time}\includegraphics[width=85mm]{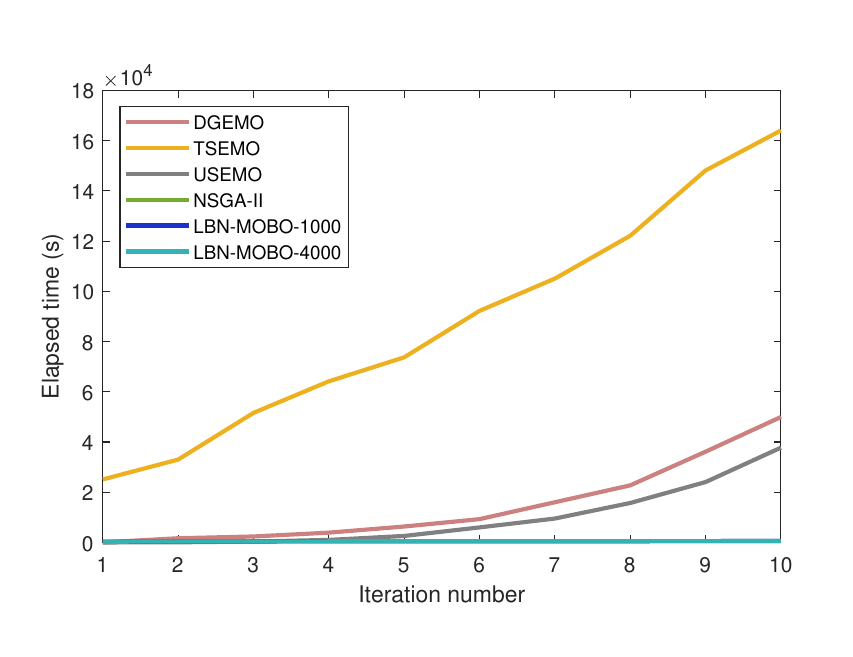}}
\subfigure[Hypervolume expansion]{\label{fig:DLTZ4_hv}\includegraphics[width=85mm]{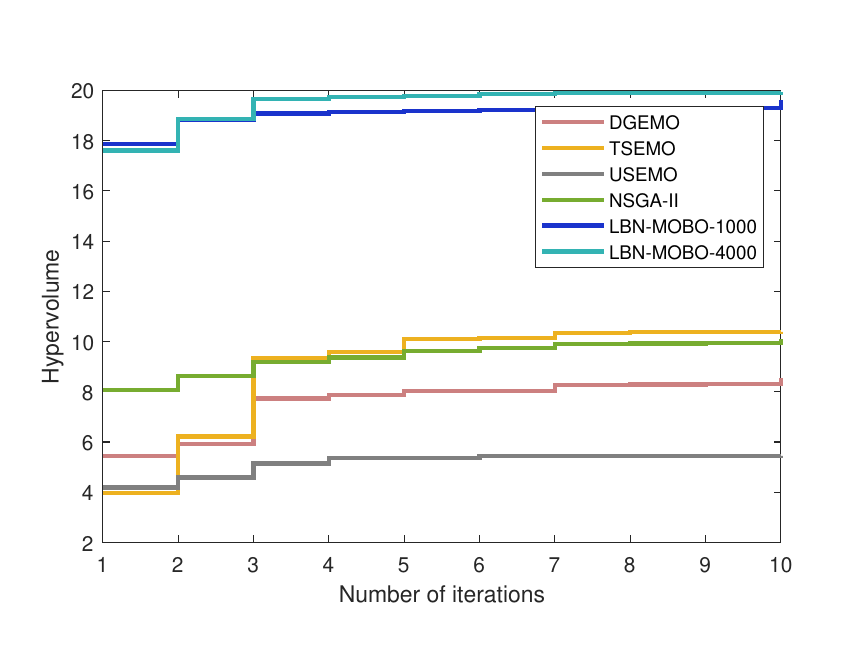}}
\subfigure[Regret analysis]{\label{fig:regrets_DTLZ4}\includegraphics[width=85mm]{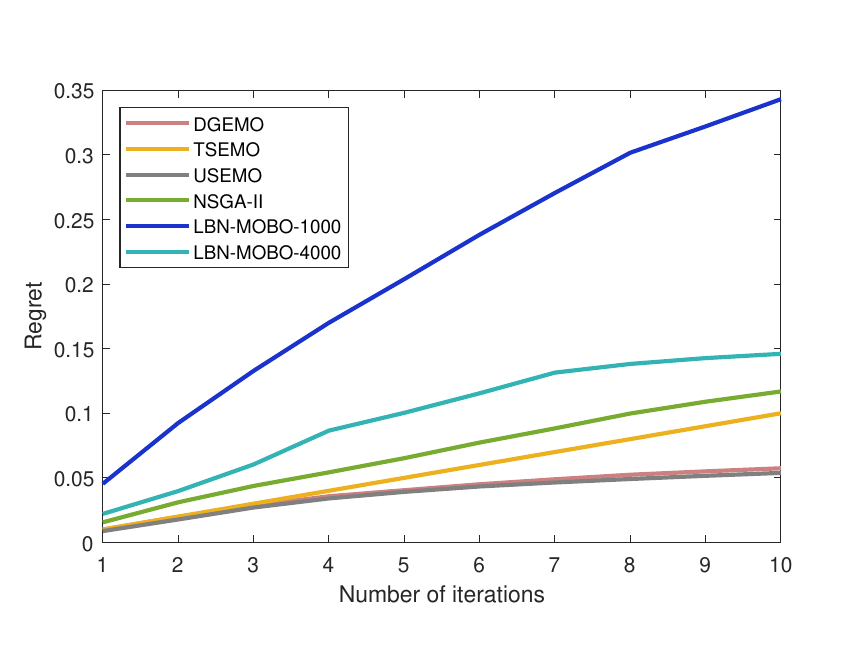}}
\caption{The optimization time, hypervolume expansion and regret analysis ofDTLZ4 experiment with 6 inputs and 3 outputs.}
\label{fig:DLTZ4_hv_time}
\end{figure}
In this section, we aim to compare the performance of our method on two problems from the DLTZ test suite \cite{deb2005scalable} with a 6-dimensional design space and a 3-dimensional performance space.
We address both the DLTZ1 and DLTZ4 problems using the same configuration described in the ZDT3 problem Section \ref{sec:ZDT_problems}.
during these experiment we have used LBN-MOBO with Deep Ensembles as the surrogate model.
Note that for a 3 dimentional output we need to solve a 6 dimentional problem while running our 2MD acquisition function.
As the dimensional of the output increases We are going to need more sample budget for our 2MD acquisition function.
As a result we solve the DTLZ4 problem using LBN-MOBO with both 1000 and 4000 sample budget.

Figures \ref{fig:DLTZ4} and \ref{fig:DLTZ4_hv} illustrates that not only LBN-MOBO has achieved the best Pareto front but also in term of computation time it is by far more scalable than the counterpart methods.
Notably, even by increasing the batch size to 4000 samples, which resulted in even better Pareto front, the computation time is still insignificant compared to the rival methods.

%

%
% \begin{figure}
%      \centering
%      \begin{subfigure}[b]
%          \centering
%                 \includegraphics[width=85mm]{figs/Time_DTLZ1.pdf}
%        \caption{Optimization time}
%          \label{fig:DLTZ1_time}
%      \end{subfigure}
%      \hfill
%      \begin{subfigure}[b]
%          \centering
%        \includegraphics[width=85mm]{figs/Progress_hv_DTLZ1.pdf}
%     \caption{Hypervolume expansion}
%     \label{fig:DLTZ1_hv}
%      \end{subfigure}
%      \hfill
%           \begin{subfigure}[b]
%          \centering
%        \includegraphics[width=85mm]{figs/regrets_DTLZ1.pdf}
%     \caption{Regret analysis}
%     \label{fig:regrets_DTLZ1}
%      \end{subfigure}
%      \hfill
%         \caption{DTLZ1 experiment with 6 inputs and 3 outputs.}
%         \label{fig:DLTZ1_hv_time}
% \end{figure}
\begin{figure}
\centering
\subfigure[Optimization time]{\label{fig:DLTZ1_time}\includegraphics[width=85mm]{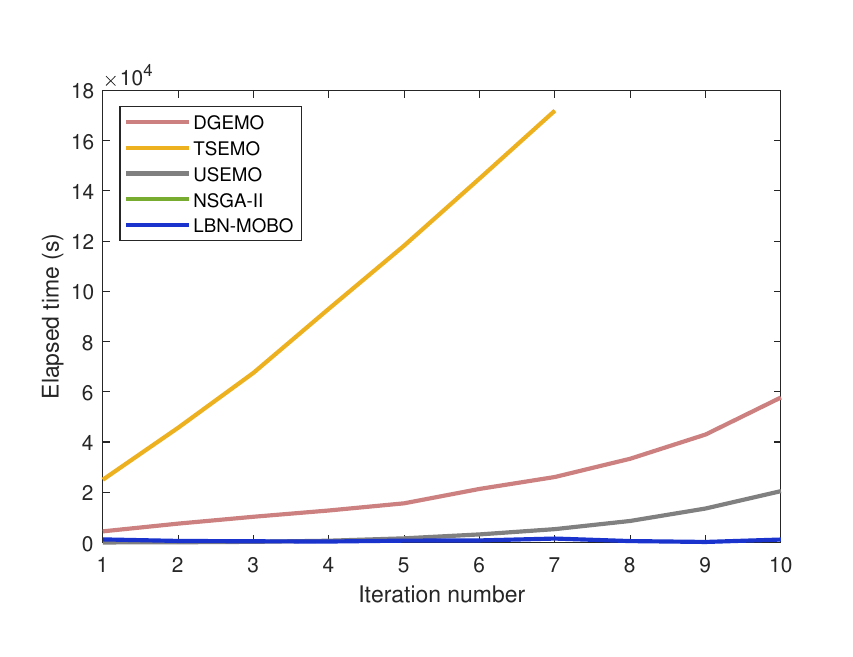}}
\subfigure[Hypervolume expansion]{\label{fig:DLTZ1_hv}\includegraphics[width=85mm]{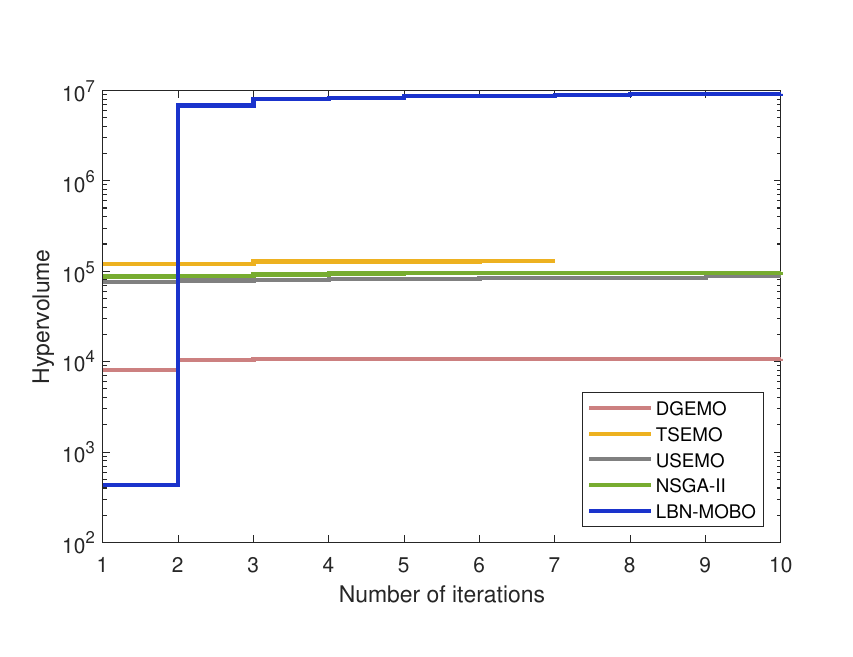}}
\subfigure[Regret analysis]{\label{fig:regrets_DTLZ1}\includegraphics[width=85mm]{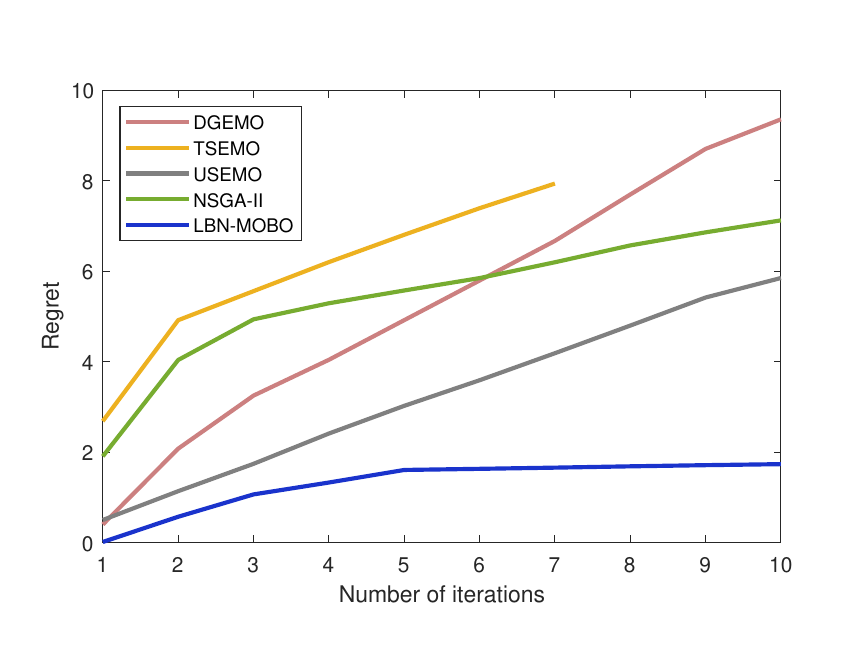}}
\caption{The optimization time, hypervolume expansion and regret analysis of DTLZ1 experiment with 6 inputs and 3 outputs.}
\label{fig:DLTZ1_hv_time}
\end{figure}

Figure \ref{fig:DLTZ1_hv} suggests that DLTZ1 problem follows the same pattern except that in this case the achieved Pareto front is orders of magnitude better than the counterpart methods. 
Likewise from Figure \ref{fig:DLTZ1_time} we understand that the computation time for LBN-MOBO is also significantly lower than the other methods.

{The regret analysis, as depicted in Figures \ref{fig:regrets_DTLZ1} and \ref{fig:regrets_DTLZ4}, corresponding to the DTLZ1 and DTLZ4 problems respectively, indicates a markedly lower regret for the LBN-MOBO approach in the DTLZ1 problem compared to its counterparts. While the regret for the DTLZ4 problem appears to increase significantly, it is noteworthy that the inherent parallelizability of LBN-MOBO allows for substantial increases in batch size. This enhancement is achieved with minimal impact on computational time, yet yields considerable improvements in terms of regret minimization and hypervolume expansion. These observations underscore the efficiency of LBN-MOBO, particularly in scenarios where batch processing capabilities can be leveraged to improve convergence.}
%%%%%%%%%%%%%%%%%%%%%%%%%%%%%%%%%%%%%%
\subsection{Real-world experiment set up} \label{sec:Experiments} 

\subsubsection{Airfoil} \label{sec:experiment_airfoil}
%
% \begin{wrapfigure}[11]{r}{0.16\textwidth}
%   \vspace{-25mm}
%   \centering
%   \includegraphics[width=0.20\textwidth]{figs/Airfoil_NFP.pdf}\\
%   \label{fig:Airfoil_NFP}
% \end{wrapfigure}
\begin{figure}[h]  % Use 'h' to suggest placement here
  \centering
  \includegraphics[width=0.20\textwidth]{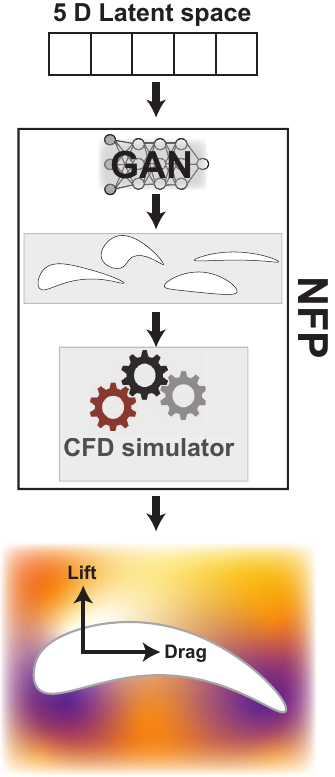}  % Adjusted width of the image
  \label{fig:Airfoil_NFP}
\end{figure}

Airfoil represents, for example, the cross-sectional shape of an airplane wing, with its performance quantified by the lift coefficient $C_{L}$ and the lift-to-drag ratio $C_{L}/C_{D}$ \cite{park2010optimal}.  

The aim of this experiment is to explore different airfoil shapes to discover the Pareto front of these performances ($C_{L}$ and $C_{L}/C_{D}$). 
Lift is the upward force that acts perpendicular to the direction of incoming airflow, primarily serving to counterbalance the weight of an aircraft or providing an upward thrust for an airfoil.
Drag is the resistance encountered by an object as it moves through a fluid. It acts in the opposite direction to the free stream flow and parallel to it. 
 
Minimizing the drag is important for maximizing the efficiency and speed of vehicles, as well as reducing fuel consumption.
In standard computational fluid dynamics (CFD) simulations, the Navier-Stokes equations are solved around the airfoil to compute $C_{L}$ and $C_{L}/C_{D}$.
We utilize the open-source software OpenFOAM for running our simulations, setting the free stream angle to 0 and length to 40 \cite{OpenFOAM,thuerey2020deep}.
The design parameters of this problem describe the shape of the airfoils.
Due to high dimensionality and complex shape constraints, we employ a specific type of Generative Adversarial Networks (GANs) to transform the complex design space into a manageable five-dimensional latent space \cite{chen2021mo}.
We assess the shapes generated by GAN using a Computational Fluid Dynamics (CFD) simulator to measure the values of $C_{L}$ and $C_{L}/C_{D}$. As such, our NFP in this problem is a combination of the GAN and the CFD simulator (inset).

\subsubsection{3D printer's color gamut} \label{sec:experiment_AB}
% \vahid{first paragraph should be halved}
A color gamut represents the range of colors that can be achieved using a specific device, such as a display or a printer \cite{wyszecki2000color}.
In this experiment, we compute the color gamut of a 3D printer by determining the Pareto front of a multi-objective optimization problem. 
A printer combines different amounts of its limited number of inks to create a range of colors.
The design parameters of this problem are the amount of available inks. 
We explore CIEa*b* color space \cite{CIE2004} which is our performance space.
CIE a* represents the color-opponent dimension of red-green, with negative values representing green and positive values representing red.
CIE b* represents the color-opponent dimension of blue-yellow, with negative values representing blue and positive values representing yellow.

Following \cite{ansari2022autoinverse} we create a printer NFP using an ensemble of 10 neural networks (not related to our ensemble surrogate). We create a complex instance of this problem where we simulate a printer NFP with 44 inks \cite{ansari2021mixed}.  All networks in the ensemble NFP are trained on 344,000 printed patches with varying ink-amount combinations and their corresponding a*b* colors. This problem has a 44 dimensional design space as the printer NFP assumes 44 inks.
%%%%%%%%%%%%%
\subsection{Complementary experiments on the real world problems}
Following the experiments in Section \ref{sec:evaluation} we compare  LBN-MOBO with NSGA-II and DGEMO on our real-world problems. 
 Other than LBN-MOBO, NSGA-II is the sole method capable of managing a batch sizes of 20,000. Additionally, we consider DGEMO due to its competitive performance in ZDT problems (Section \ref{sec:ZDT_problems}), although we must limit its batch size to 1,000 and restrict it to 5 to 6 iterations due to prohibitive run time.

\subsubsection{Evaluation of 44-ink printer gamut experiment using NSGA II and DGEMO} \label{sec:complementary_eval_printer}
\begin{figure}
\centering
\subfigure[Hypervolume evolution]{\label{fig:Progress_hv_AB}\includegraphics[width=85mm]{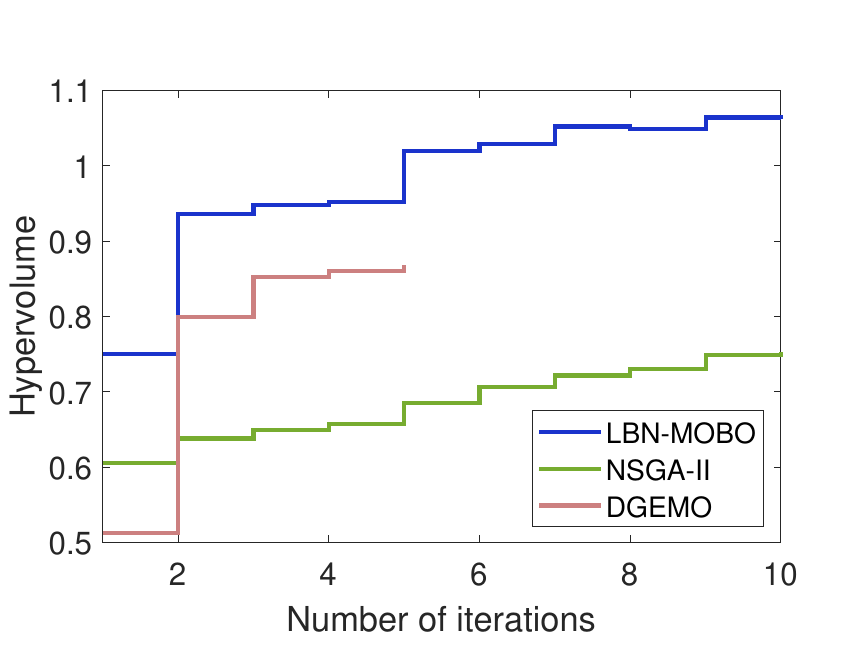}}
\subfigure[Piecewise linear color gamut at iteration 10]{\label{fig:44_ink_gamut}\includegraphics[width=85mm]{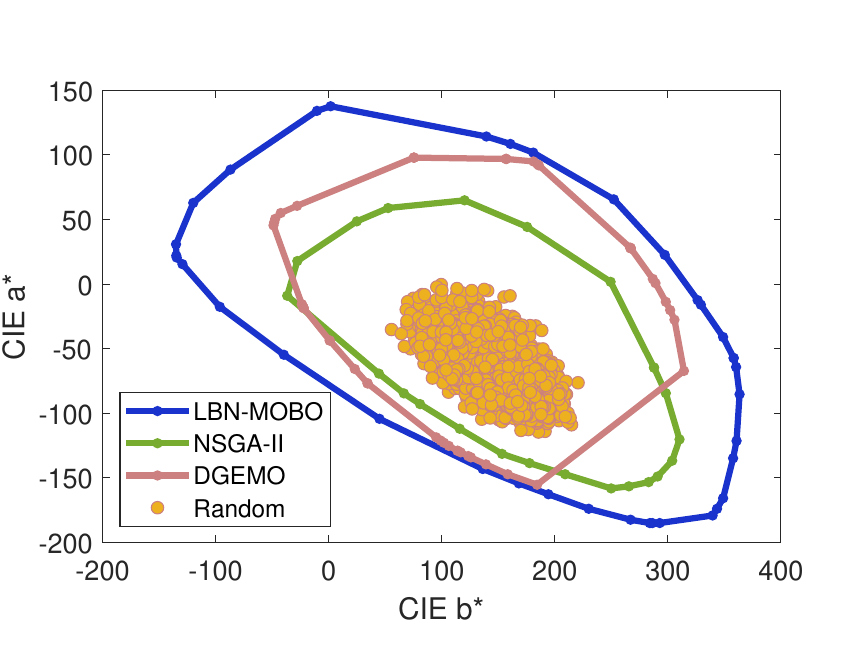}}
\caption{The hypervolume evolution and gamut of the 44-ink printer calculated by different methods.}
\label{fig:hv}
\end{figure}

%%%%%% 44 ink Printer %%%%%%
For the task of exploring the 44-ink color gamut, we initialize LBN-MOBO with 10,000 samples, and each subsequent iteration processes a batch size of 20,000 samples. Given the high dimensionality of the design space, this problem poses a significant challenge to many optimization algorithms, making it a fascinating experimental case. The performance space in this experiment is the 2 dimensional a*b* color space.
Figure \ref{fig:Progress_hv_AB} graphically depicts the accelerated increase in hypervolume of the color gamut when using LBN-MOBO.
Also, final gamut estimation for NSGA-II and LBN-MOBO after 10 iterations, and DGEMO after 5 iterations, is depicted in Figure \ref{fig:44_ink_gamut}, showing a significanlty larger estimated gamut by LBN-MOBO. 
%
%%

% \begin{figure}
%      \centering
%      \begin{subfigure}[b]
%          \centering

%                 \includegraphics[width=85mm]{figs/Progress_hv_AB.pdf}
%        \caption{Hypervolume evolution}
%          \label{fig:Progress_hv_AB}
%      \end{subfigure}
%      \hfill
%      \begin{subfigure}[b]
%          \centering
%        \includegraphics[width=85mm]{figs/Comparison_AB_44ink.pdf}
%     \caption{Piecewise linear color gamut at iteration 10}
%     \label{fig:44_ink_gamut}
%      \end{subfigure}
%      \hfill
%         \caption{The hypervolume evolution and gamut of the 44-ink printer calculated by different methods.}
%         \label{fig:hv}
% \end{figure}

%%%%% Airfoil%%%%%%
\subsubsection{Evaluation of Airfoil experiment using NSGA II and DGEMO} \label{sec:complementary_eval_airfoil}
% \begin{figure}
%      \centering
%      \begin{subfigure}[b]
%     \centering
%      \includegraphics[width=85mm]{figs/Progress_hv_airfoil.pdf}
%      \caption{Hypervolume evolution}
%     \label{fig:Progress_hv_airfoil}
%      \end{subfigure}
%      \hfill
%      \begin{subfigure}[b]
%     \centering
%         \includegraphics[width=85mm]{figs/comparison_airfoil.pdf}
%     \caption{Piecewise linear Pareto front at iteration 5}
%     \label{fig:comparison_airfoil}
%      \end{subfigure}
%      \hfill
%           \begin{subfigure}[b]
%     \centering
%         \includegraphics[width=85mm]{figs/regrets_airfoil.pdf}
%     \caption{Regret analysis}
%     \label{fig:regrets_airfoil}
%      \end{subfigure}
%      \hfill
%         \caption{The hypervolume evolution and the Pareto front of the airfoil problem of LBN-MOBO and NSGA-II with equal batch sizes. }
%         \label{fig:three graphs}
% \end{figure}
\begin{figure}
\centering
\subfigure[Hypervolume evolution]{\label{fig:Progress_hv_airfoil}\includegraphics[width=85mm]{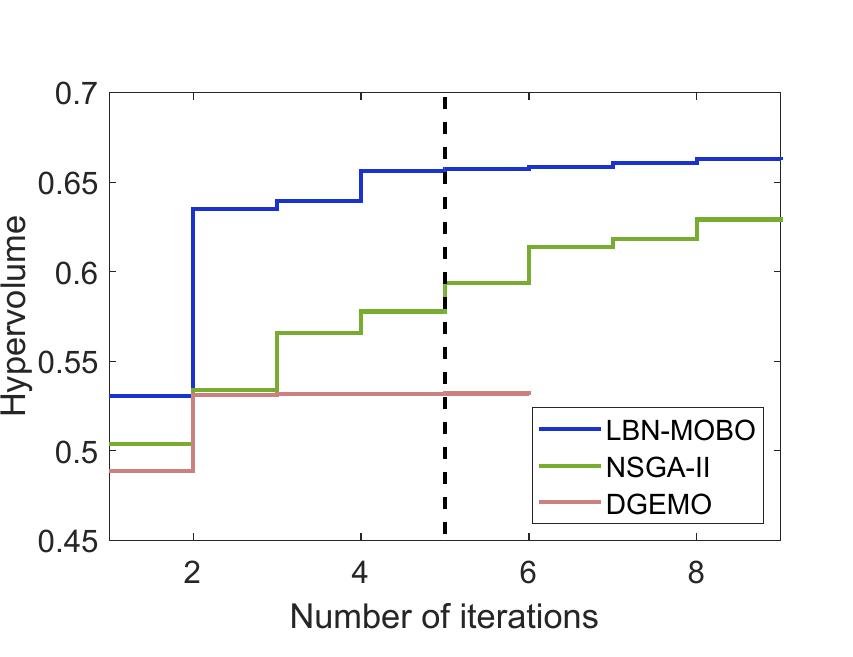}}
\subfigure[Piecewise linear Pareto front at iteration 5]{\label{fig:comparison_airfoil}\includegraphics[width=85mm]{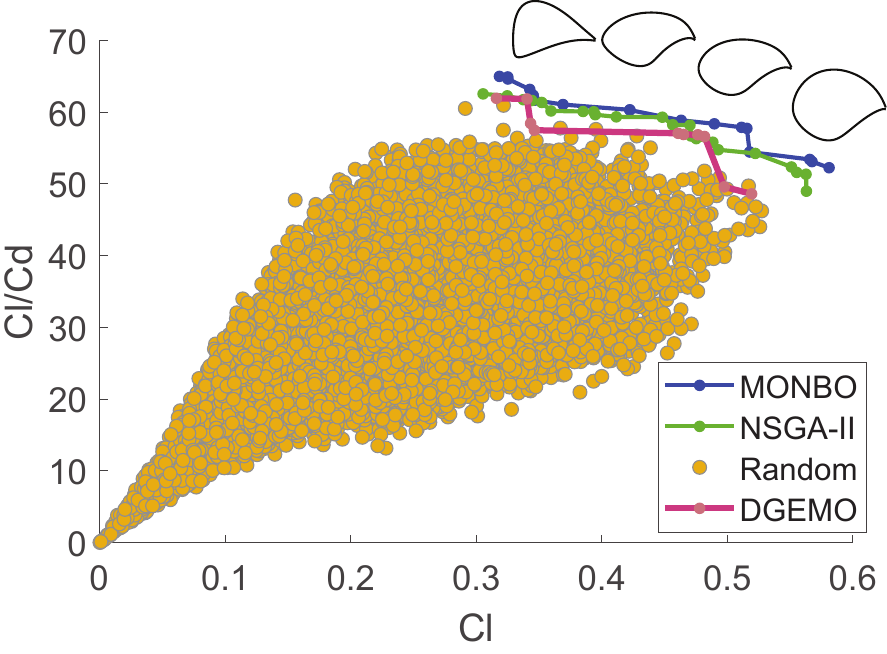}}
\subfigure[Regret analysis]{\label{fig:regrets_airfoil}\includegraphics[width=85mm]{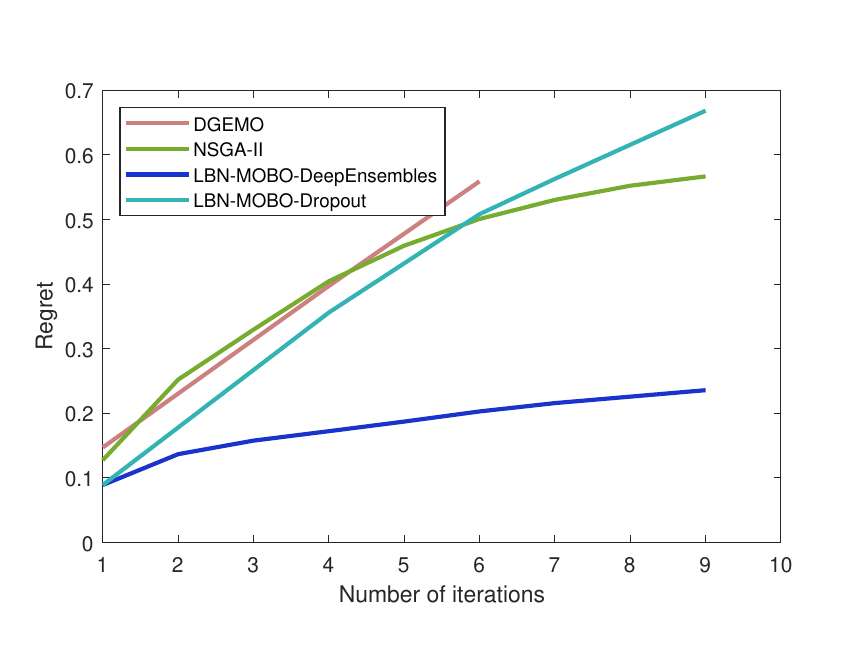}}
\caption{The hypervolume evolution, the Pareto front, and regret analysis of the airfoil problem. LBN-MOBO and NSGA-II have equal batch sizes, however DGEMO has a batch size of 1000 due to its computational bottleneck.}
\label{fig:three graphs}
\end{figure}

Figure~\ref{fig:three graphs} showcases a comparison between NSGA-II, DGEMO, and LBN-MOBO for the airfoil problem. This experiment has a significantly more complex NFP since the relationship between the design and performance space is highly complex as we map the latent code of a GAN to aerodynamic properties. We start LBN-MOBO and NSGA-II with 15,000 samples, and each iteration runs with a batch size of 15,000 all simulated by OpenFOAM, a high-fidelity CFD solver \cite{OpenFOAM}. 

Once again, as depicted in Figure~\ref{fig:Progress_hv_airfoil}, LBN-MOBO discovers a superior Pareto front in a remarkably small number of iterations. 
% than the other methods. thanks to its large batch and its high-capacity surrogate as well as its exploratory acquisition function, 
Although with many more iterations, NSGA-II also reaches an acceptable Pareto front, this is most likely due to the use of a large batch size for each iteration in a comparatively smaller design space (five dimensions). In contrast, DGEMO's performance significantly deteriorates. This likely stems from its reliance on the precise estimation of the gradient and Hessian of the NFP through the surrogate model. This task becomes increasingly difficult as the complexity of the NFP increases.

{Figure \ref{fig:regrets_airfoil} illustrates the cumulative regret for various optimization methods applied to this problem. Conducting regret analysis in real-world scenarios presents additional complexities, primarily due to the lack of a known Pareto optimal solution which is essential for calculating the regret as defined in equation \ref{eq:batch_MO_regret}.
In our experimental setup, the best-discovered Pareto front is assumed to be the Pareto optimal solution. This assumption forms the basis for calculating the regret distances.
Analysis of the data presented in Figure \ref{fig:regrets_airfoil} reveals that LBN-MOBO demonstrates a significantly lower increase in cumulative regret when compared to its counterparts. This outcome highlights the effectiveness of LBN-MOBO in minimizing regret, thereby indicating its superior performance in navigating towards optimal solutions in real-world problem settings.}

%%%%%%%%%%%%%%%%%%%%%%%%%%%%%%%%%%%%%%%%%%%%%%%%%%%%%%%%%%%%%%%%%%%%%%%
\subsection{Printer color gamut for 8 ink} 
\begin{figure*}
    \centering
        
        \includegraphics[width=\textwidth]{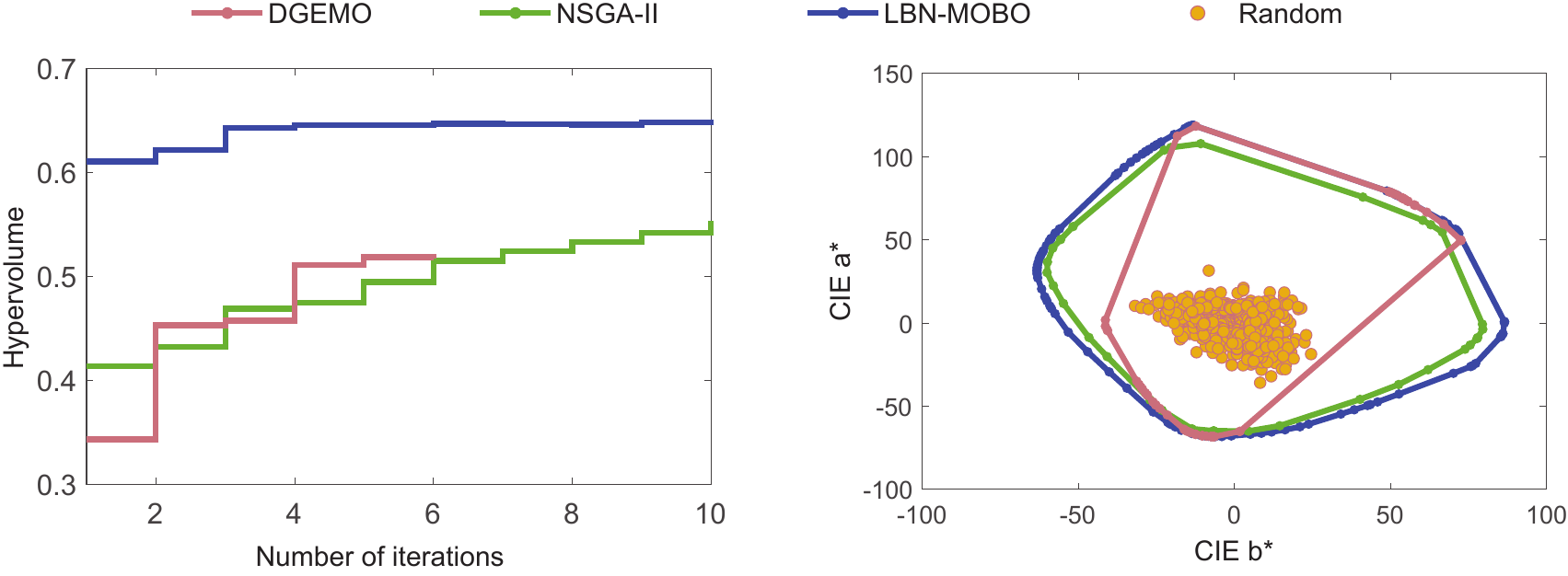}
    \caption{The Hyper volume and gamut of the 8-ink printer. NSGA-II and LBN-MOBO have the same batch size while DGEMO has a batch size of 1000 due to its lack of scalability.}
    \label{fig:8-ink-AB}
\end{figure*}
Figure \ref{fig:8-ink-AB} displays the results of the 8-ink printer color gamut problem, which has a setup that is nearly identical to the 44-ink problem, with smaller number of available inks and as a result smaller design space.
The advantage of this NFP, is in its ability in accurately mimicking the behavior of the Epson printer from which the data set is derived.
% \cite{ansari2020mixed}
%

As observed, both NSGA-II and DGEMO exhibit results that are closer to LBN-MOBO, which could be attributed to the lower dimensionality of the problem.

%%%%%%%%%%%%%%%%%%%%%%%%%%%%%%%%%%%%%%%%%%%%%%%%%%%%%%%%%%%%%%%

\section{Implementation Details} \label{sec:implementation_detail}

\subsection{Deep Ensembles}
For our surrogate model, building on the methodology proposed in \cite{ansari2022autoinverse}, we have constructed Deep Ensembles, utilizing a diverse collection of activation functions, to enhance the precision of epistemic uncertainty quantification. This enhancement serves as a cornerstone for ensuring the robust operation of the remaining procedures.

In our implementation, the Deep Ensembles comprises ten sub-networks, each employing a specific activation function as outlined below:
\begin{itemize}
\item Tanh  $\times 2$ \cite{lecun2002efficient}
\item ReLU  $\times 2$ \cite{nair2010rectified}
\item CELU  $\times 2$ \cite{barron2017continuously}
\item LeakyReLU  $\times 2$ \cite{maas2013rectifier}
\item ELU \cite{clevert2015fast}
\item Hardswish \cite{howard2019searching}
\end{itemize}

For the ZDT3 and printer's color gamut networks, we employ a three-layer architecture, with neuron configurations of 100, 50, and 100 per layer respectively.
To conduct our analysis on the rival methods, we utilized the  pymoo library \cite{pymoo} and DGEMO source code \cite{konakovic2020diversity}.

Given the complexity inherent to the airfoil problem, it necessitates a more intricately designed network. We configured this network with four hidden layers, containing 150, 200, 200, and 150 neurons, respectively.
In the Native Forwared Process (NFP) of the Airfoil design, i.e., the open source fluid simulator OpenFoam, we have observed that sampling the latent space of the GAN near 0 and, in general, below 0.1 occasionally leads to invalid designs.
This occurrence can introduce instability for the optimizers. To address this issue, we have imposed a limitation on the GAN latent space, restricting it between 0.1 and 1 to ensure the generation of valid designs.

Over the course of the LBN-MOBO iterations, data accumulation intensifies. Although it is feasible to progressively enlarge the batch size to maintain a constant total training time, we have chosen to keep the batch sizes fixed due to the minimal increment in training time relative to competing methods. Specifically, we employed a batch size of 20 for the airfoil problem, 10 for ZDT3, and 100 for the printer's color gamut.

All networks underwent a training period spanning 60 epochs.
\subsection{MC dropout} \label{sec:mcdropout}

As a general rule of thumb, we opt for wider architectures compared to those in the Deep Ensembles network. This approach is based on the consideration that, since a dropout layer is utilized in every training instance, some of the perceptrons are deactivated, leading to a somewhat narrower sub-network in the trained model. The dropout ratio for all the models is set at 0.05, and ReLU is employed as the activation function. To compute the epistemic uncertainty for every inference, each network is queried 100 times.

For the ZDT3 problem, a network consisting of four fully connected layers with 100, 50, 50, and 100 perceptrons respectively, is trained for 100 epochs. The batch size in this instance is set at 5.

Addressing the printer's color gamut problem, we use a network configuration with three layers, containing 200, 100, and 200 perceptrons. This model is trained with a batch size of 100 for 80 epochs.

Given the Airfoil problem's increased complexity, a network comprising four layers with 300, 400, 400, and 300 perceptrons is employed for modeling. The batch size designated for this problem is 20, and the network is trained for 160 epochs.

\subsection{Hardware configuration} \label{sec:hardware}
We leveraged a parallel compute cluster consisting of GPUs and CPUs for the simultaneous training of the network and computation of the acquisition function.

The GPU units within our cluster comprise two models: the NVIDIA Tesla A100, the NVIDIA Tesla A40, and NVIDIA Tesla A16.
The GPU units have a memory up to 64GB.

The CPU units in the cluster are AMD EPYC 7702 64-Core Processor.
\section{Complementary Discussions}
{\subsection{Refining Deep Ensembles: focusing solely on epistemic uncertainty}
\label{sec:why_no_aleatoric}
In this study, we interpret epistemic uncertainty as the uncertainty stemming from incomplete knowledge about a process, which can be mitigated through additional data collection. Conversely, aleatoric uncertainty, indicative of inherent randomness, is irreducible. Our work operates under the premise that the Native Forward Processes (NFPs) involved exhibit minimal noise. 
Given the negligible impact of aleatoric uncertainty in scenarios with low noise, we have adapted Deep Ensembles to focus solely on epistemic uncertainty, which is crucial for exploration phases. This approach streamlines the model training process, as it removes the need for training separate networks for variance estimation associated with aleatoric uncertainty, thereby enhancing stability and simplicity in the training stage \cite{nix1994estimating, seitzer2022pitfalls}.

{To further support our intuition we repeat the experiments from Sections \ref{sec:ZDT3} and \ref{sec:ZDT1_ZDT2} this time we use the original Deep Ensembles model including aleatoric uncertainty. Figure \ref{fig:LBN_MOBO_aleatoric} indicates that incorporating aleatoric uncertainty does not affect the exploration process in a beneficial manner, as showcased by the ZDT1 results, and may even delay the convergence, as observed in the ZDT2 and ZDT3 experiments.}

\begin{figure}[h]
    \centering
    \includegraphics[width=0.4\textwidth]{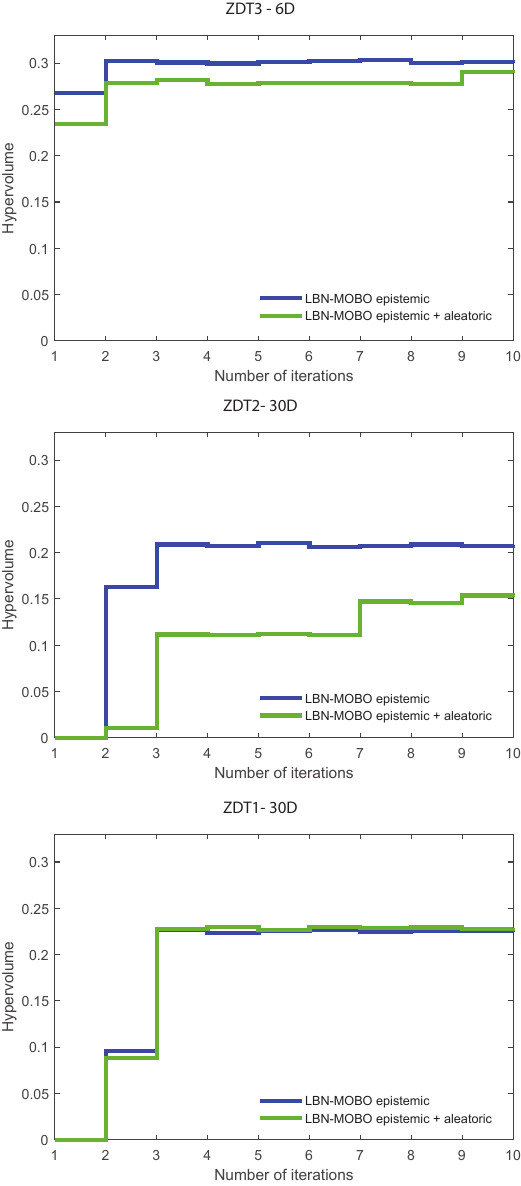}
    \caption{We ran the experiments from Sections \ref{sec:ZDT3} again, but this time we included calculations for aleatoric uncertainty as well as epistemic uncertainty. As evident from the results using the aleatoric uncertainty during the exploration not only does not improve the results but in some cases it also delays the convergence. It is mainly due to the nature of the aleatoric uncertainty that is an indication of the existence of the irreducible noise and is not a good measure for exploitative purposes. The batch size in all the experiments is fixed to 1000 samples.}
    \label{fig:LBN_MOBO_aleatoric}
\end{figure}

\subsection{Similarity and distinctions between LBN-MOBO and its counterparts}
\label{sec:similarity_distinction}
\paragraph{USeMO}
\textbf{USeMO} \cite{belakaria2020uncertainty} shares some similarity with \textbf{LBN-MOBO}, however, besides its limitation in handling large batches, its utilization of uncertainty information is much more limited than ours.
\textbf{USeMO} finds the Pareto front on their surrogate function by running the NSGA-II \cite{deb2002fast}. Note that they do not optimize for uncertainty in a simultaneous manner as it happens in our \textbf{2\textit{M}D} acquisition. Instead they use uncertainty in a sequential manner to choose the most promising candidates among the ones already calculated by the NSGA-II.
This approach cannot account for the samples that are not Pareto dominant according to performance predictions but have high uncertainty. 
The results in Figures \ref{fig:zdt3_pareto} , \ref{fig:ZDT1_2}, \ref{fig:DLTZ4_hv_time}, \ref{fig:DLTZ1_hv_time} in the Appendix also confirm that \textbf{USeMO} is not very successful in recovering a diverse Pareto front.

\paragraph{TSEMO}
Similar to \textbf{USeMO}, \textbf{TSEMO} \cite{bradford2018efficient} utilizes NSGA-II for the calculation of the approximated Pareto set and Pareto front on the computationally inexpensive surrogates.
What brings \textbf{TSEMO} closer to our approach is their utilization of Thompson sampling \cite{thompson1933likelihood} to exploit or explore the black box function, guided by the uncertainty information obtained from Gaussian process surrogates.
\paragraph{UCB}
{Upper confidence bound (\textbf{UCB})} and \textbf{2$M$D} Pareto front exploit the uncertainties in different ways. \textbf{UCB} works based on a weighted sum of uncertainty and prediction. In \textbf{UCB}, tuning is important as it regulates the trade-off between exploration and exploitation. Given the magnitude difference between prediction and uncertainty, one needs to tune this hyperparameter. Moreover, as we progress through the optimization the need for exploration or exploitation might change and as a result we might need to constantly adapt this hyperparameter or use a smart scheduler. In the case of multi-objective optimization, the number of tuning parameters increases as we need to tune and find the right balance in exploring different objectives in the presence of their uncertainties.

In the \textbf{2\textit{M}D} acquisition function, however, the uncertainties and predictions remain independent objectives and are jointly optimized to generate a Pareto front of viable candidates (Section \ref{sec:acquisition} and Equation \ref{eq:acquisition}). Since each objective is optimized jointly with other objectives without any form of summation we do not need to worry about the scale differences and \textbf{LBN-MOBO} does not introduce extra hyperparameters.
\stopcontents[appendices]

\end{document}